\documentclass[letterpaper]{article} 
\usepackage{aaai2026}  
\usepackage{times}  
\usepackage{helvet}  
\usepackage{courier}  
\usepackage[hyphens]{url}  
\usepackage{graphicx} 
\usepackage{natbib}  
\usepackage{caption} 
\usepackage{algorithmic}
\usepackage{booktabs}
\usepackage{subcaption}
\usepackage{xcolor}  
\usepackage[ruled,linesnumbered]{algorithm2e}
\usepackage{amsmath}
\usepackage{makecell}
\usepackage{cleveref}
\usepackage {colortbl}
\usepackage{float}
\usepackage{amssymb}
\usepackage{multirow}
\usepackage{tabularx}

\usepackage{newfloat}
\usepackage{listings}
\DeclareCaptionStyle{ruled}{labelfont=normalfont,labelsep=colon,strut=off} 
\floatstyle{ruled}
\newfloat{listing}{tb}{lst}{}
\floatname{listing}{Listing}
\newcommand{\ours}{COVR}
\Crefname{figure}{Fig.}{Figs.}
\Crefname{table}{Table}{Tables}
\crefname{equation}{Eq.}{Eqs.}
\crefname{section}{Sec.}{Secs.}
\crefname{algorithm}{Alg.}{Algs.}

\title{COVR: Collaborative Optimization of VLMs and RL Agent for Visual-Based Control}
\author{
    Canming Xia\textsuperscript{\rm 1,2}, Peixi Peng\textsuperscript{\rm 3,2}\thanks{
    Corresponding authors.
    }, Guang Tan\textsuperscript{\rm 1}\footnotemark[1], 
    Zhan Su\textsuperscript{\rm 3}, 
    Haoran Xu\textsuperscript{\rm 1,2}, \\
    Zhenxian Liu\textsuperscript{\rm 4}, 
    Luntong Li\textsuperscript{\rm 2} 
}

\affiliations{
    \textsuperscript{\rm 1} School of Intelligent Systems Engineering, Shenzhen Campus of Sun Yat-sen University, China \quad \\
    \textsuperscript{\rm 2} Peng Cheng Laboratory, China \quad \\
    \textsuperscript{\rm 3} School of Electronic and Computer Engineering, Shenzhen Graduate School, Peking University, China \\
    \textsuperscript{\rm 4} National Engineering Research Center of Visual Technology, School of Computer Science, Peking University, China
    \\
    xiacm@mail2.sysu.edu.cn, pxpeng@pku.edu.cn, tanguang@mail.sysu.edu.cn
}

\usepackage{bibentry}

\begin{document}

\maketitle

\begin{abstract}
Visual reinforcement learning (RL) suffers from poor sample efficiency due to high-dimensional observations in complex tasks. While existing works have shown that vision-language models (VLMs) can assist RL, they often focus on knowledge distillation from the VLM to RL, overlooking the potential of RL-generated interaction data to enhance the VLM. To address this, we propose \ours, a collaborative optimization framework that enables the mutual enhancement of the VLM and RL policies. Specifically, \ours\ fine-tunes the VLM with RL-generated data to enhance the semantic reasoning ability consistent with the target task, and uses the enhanced VLM to further guide policy learning via action priors. To improve fine-tuning efficiency, we introduce two key modules: (1) an Exploration-Driven Dynamic Filter module that preserves valuable exploration samples using adaptive thresholds based on the degree of exploration, and (2) a Return-Aware Adaptive Loss Weight module that improves the stability of training by quantifying the inconsistency of sampling actions via return signals of RL. We further design a progressive fine-tuning strategy to reduce resource consumption. Extensive experiments show that \ours\ achieves strong performance across various challenging visual control tasks. 

\end{abstract}



\section{Introduction}
\begin{figure}[!t] 
    \begin{subfigure}{\linewidth}
        \includegraphics[width=1.0\linewidth]{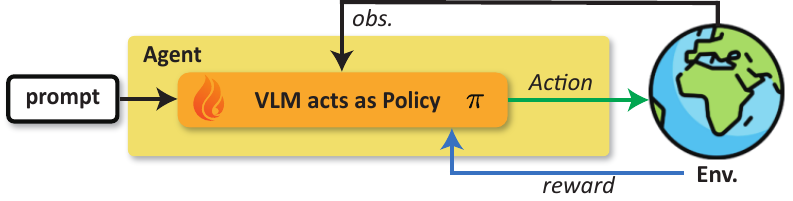}
        \caption{Fine-tuning a VLM into the policy network.}
    \end{subfigure}
    \begin{subfigure}{\linewidth}
        \includegraphics[width=1.0\linewidth]{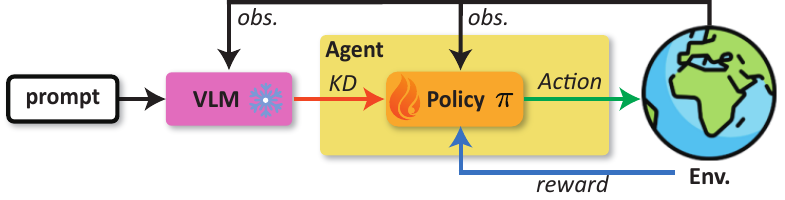}
        \caption{Knowledge distillation from a VLM to the policy network.}
    \end{subfigure}
    \begin{subfigure}{\linewidth}
        \includegraphics[width=1.0\linewidth]{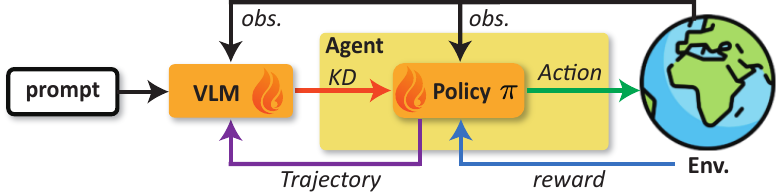}
        \caption{Our collaborative optimization framework that integrates a VLM with RL. It iteratively enhances the capabilities of VLMs and RL agents.}
    \end{subfigure}
    \caption{Comparison of VLM-adapted and VLM-assisted policy learners.}
    \label{Fig:1}
    
\end{figure}

Visual reinforcement learning (RL) has emerged as a critical approach for intelligent systems to cope with complex tasks such as robotic control~\cite{fu2024furl}, autonomous driving~\cite{zhang2020learning,xu2024dmr}, and game simulation~\cite{liu2025visual, laskin2020reinforcement}. A fundamental challenge in visual RL is the inefficiency of exploration under high-dimensional visual inputs, where invalid interactions and complex state spaces complicates policy learning. Recent advancements in large-scale models (LMs), including large language models and vision-language models (VLMs), have shown potential in guiding decision making~\cite{shinn2023reflexion, wang2024describe, pan2024vlp, dalal2024plan}. Existing works have thus explored integrating LMs into RL to enhance learning performance~\cite{lee2025sample, zhou2024large, xu2025vlms}.

Existing LM-assisted methods generally fall in two categories. One is to treat the LM as a fixed feature extractor integrated into the policy module~\cite{paischer2023semantic, fu2024furl, guo2025improving}, or directly fine-tune the LM as the policy network~\cite{zhai2024fine, tan2024true, wei2025gtr}, as depicted in \Cref{Fig:1}(a). These approaches may suffer from high computational overhead and deployment costs for online inference. Another line of work aims to transfer the intrinsic knowledge of a frozen LM to the policy network via knowledge distillation (KD)~\cite{lee2025sample, zhou2024large, xu2025tell, chen2024rlingua}, see an illustration in \Cref{Fig:1}(b). These approaches can improve policy learning using priors and allow faster inference, but may yield suboptimal performance when the VLM is trained with limited domain-specific data. This is due to the frozen VLM that potentially propagates inaccurate reasoning results, leading to negative impact on policy learning. 

We argue that VLMs and visual RL agents possess highly complementary strengths, and thus, a principled integration of the two is both natural and beneficial:  
(1) \textbf{VLM improves RL via Prior Guidance}. VLMs can extract rich semantic priors from images and provide direct, task-specific policy guidance to accelerate the training of visual RL models. In contrast to indirect methods that rely on reward shaping~\cite{xie2024text2reward, wang2024rl} or task decomposition~\cite{dalal2024plan, shukla2024lgts}, this approach enables more stable and efficient policy learning by leveraging explicit semantic reasoning from VLMs. 
(2) \textbf{RL enhances VLM via Specialized Experience}. RL is able to discover high-quality state-action pairs in specific scenarios, despite the challenges of coping with high-dimensional state space in real-world tasks. By exploiting these RL-derived trajectories to fine-tune the VLM, we effectively encode domain-specific strategies into the VLM. This enables task-oriented prior generation that could generalize to unseen states.


Building on these insights, we propose a novel approach for VLM-assisted RL, implemented through a collaborative optimization framework called \ours, as illustrated in \Cref{Fig:1}(c). It comprises two key components: 
(1) Knowledge Enhancement of RL-Tuned VLM: The VLM is fine-tuned using interaction trajectories collected from the RL agent. This process aligns the VLM’s ability with task-specific semantics, enhancing its decision relevance. 
(2) VLM-Guided Policy Learning: The fine-tuned VLM provides generalizable action priors that guide the RL policy. These priors accelerate learning by offering more informative gradients. To effectively realize \ours, we address two key challenges: 
\begin{itemize}
    \item How to select high-quality trajectory samples? We design an \textit{Exploration-Driven Dynamic Filter} (EDDF) module, which receives state-action pairs generated during RL interactions, along with their associated return computed from accumulated episode rewards. The module dynamically determines the thresholds based on the agent's exploration level to select high-quality samples; 
    \item How to alleviate the impact of action inconsistency in RL? During training, RL may yield distinct high-reward actions for similar observations, resulting in inconsistent signals for the VLM and potentially hindering policy convergence. We introduce the \textit{Return-Aware Adaptive Loss Weight} (RALW) module, which dynamically adjusts the loss weights according to return values. This mechanism enables the model to prioritize high-return samples while preserving its original capabilities on low-return examples. 
\end{itemize}

Finally, we introduce an {\em Adaptive Progressive Fine-Tuning} strategy that gradually reduces the frequency of VLM updates as the RL policy converges, thereby improving overall training efficiency.

In summary, the contributions of this work are fourfold: 
\begin{itemize}
    \item A novel VLM-assisted RL method based on collaborative optimization, which enables visual RL policies to achieve improved performance even when the assisting VLM has limited capabilities. 
    \item An EDDF module that dynamically selects trajectory samples based on the exploration degree of RL to enhance task-specific knowledge acquisition in VLMs. 
    \item A RALW module that distinguishes inconsistent actions under similar observations by adaptively adjusting loss weights based on return signals, leading to more effective learning of the VLM. 
    \item Extensive experiments on the CARLA~\cite{dosovitskiy2017carla} and DMControl~\cite{tassa2018deepmind} validating the superiority of our method.
\end{itemize}

\section{Related Work}
\paragraph{LM-assisted RL.}
Most methods that integrate LMs into RL can be categorized into three classes: 
(1) \textbf{LMs as agents}. In this approach, LMs serve as core components of the policy. It is further divided into parametric and non-parametric methods. Parametric methods fine-tune the LM to generate task-specific actions, enabling strong adaptation to downstream tasks~\cite{zhai2024fine, tan2024true, wei2025gtr}, while non-parametric methods utilize a frozen LM to extract rich semantic priors for decision-making~\cite{zhou2024large, shinn2023reflexion}, with wide applications in robotic manipulation~\cite{chen2024rlingua} and agent collaboration~\cite{xu2024language}. However, parametric methods face deployment challenges, while non-parametric ones struggle with long-term planning. Our method addresses these issues via knowledge distillation and enhances sequential planning in continuous RL tasks. 
(2) \textbf{LMs as planners}. The LM decomposes tasks into sub-goals, generating plans either upfront~\cite{tang2023saytap, shukla2024lgts} or incrementally~\cite{lee2025sample,zhou2024large}. Yet, these rely heavily on the LM' reasoning, which our framework mitigates through return-aware filtering. 
(3) \textbf{LMs as reward shapers}. In this setting, the LMs assists RL by modeling reward signals, either by generating executable reward functions~\cite{xie2024text2reward} or by producing scalar reward estimates~\cite{wang2024rl, wang2025prefclm}. Despite their potential, these approaches often fail to capture real-world complexity and incur high trial-and-error costs. In contrast, our method provides direct policy guidance, significantly shortening the path from knowledge to effective action.

\paragraph{Training LMs with RL.}
RL has been extensively utilized to elicit and refine capabilities in LMs. Among these approaches, RL from human feedback (RLHF) trains reward models using human-labeled data before optimizing the policy~\cite{team2024gemini}. Alternatively, methods incorporating human preference data shape the reward function to align model behavior with human expectations~\cite{rafailov2023direct}. In contrast, our method performs RL fine-tuning based on environmental rewards. 
Recent advances in process reward models~\cite{lightman2023let}, search algorithms~\cite{xin2024deepseek}, and GRPO~\cite{shao2024deepseekmath} have achieved notable progress in specialized tasks like mathematical reasoning. However, our focus is on sequential decision-making for goal-directed behaviors in interactive environments. By integrating visual input and visual-language reasoning, we have enhanced the performance of RL in multiple tasks. While recent work~\cite{waite2025rls3} leverages RL to generate data for improving the VLM, it does not incorporate this knowledge back into the RL policy, resulting in high exploration costs. Conversely, our method is designed to enhance policy learning within RL by feeding knowledge back into the policy, thereby reducing exploration costs.

\section{Preliminary}
\paragraph{Process of Visual RL.}
The visual RL task can be modeled as a Markov Decision Process (MDP), represented by the tuple $(\mathcal{O},\mathcal{S},\mathcal{A},\mathcal{T},\mathcal{R},\gamma)$. In this formulation, $o_t\in\mathcal{O}$ represents the raw visual input observed at time step $t$, while $s_{t} \in \mathcal{S}$ denotes the corresponding state features extracted from $o_t$. The set $\mathcal{A}$ defines the space of actions, $\mathcal{R}$ is the reward function, and $\gamma \in [0,1]$ serves as the discount factor. At each time step, the agent receives the visual observation $o_t$ and generates an action $a_t\in\mathcal{A}$ according to the policy $\pi$. Upon executing the action, the environment returns a scalar reward $r_{t+1} \sim \mathcal{R}(s_{t},a_{t})$, the next visual observation $o_{t+1}$, and the game done flag.

\paragraph{Soft Actor Critic.}
Our method is built upon the Soft Actor-Critic (SAC)~\cite{haarnoja2018soft1, haarnoja2018soft2}, which alternately optimizes a actor network $\pi_{\theta}(\cdot)$ and a critic network $Q_{\phi}(\cdot)$. SAC aims to maximize the expected long-term reward while promoting exploration through an entropy regularization term weighted by $\alpha$:
\begin{equation}
\mathcal{L}_{\pi}=-\mathbb{E}_{a_{t}\sim\pi_{\theta}}\left[Q_{\phi}(s_{t},a_{t})-\alpha\log\pi_{\theta}(a_{t}|s_{t})\right].
\label{eq:1}
\end{equation}

The parameters of $Q_{\phi}(\cdot)$ are updated using the Bellman backup target, formulated as:
\begin{equation}
\mathcal{L}_Q=\mathbb{E}_{(s_t,a_t)\sim\mathcal{D}}\left[(Q_{\phi}(s_t,a_t)-(r_t+\gamma V(s_{t+1})])\right)^2,
\label{eq:2}
\end{equation}
where $\mathcal{D}$ denotes the experience replay buffer. The soft state value function $V(s_{t+1})$ is computed as:
\begin{equation}
V(s_{t+1})=\mathbb{E}_{\tilde{a}\sim\pi_{\theta}}\left[\bar{Q}_{\phi}(s_{t+1},\tilde{a})-\alpha\log\pi_{\theta}(\tilde{a}|s_{t+1})\right],
\label{eq:3}
\end{equation}
where $\bar{Q}_{\phi}(\cdot)$ representing the target critic network, which is maintained as an exponential moving average of $Q_{\phi}(\cdot)$, and $\tilde{a}$ is the next action sampled from $\pi_{\theta}(\cdot)$.

\section{Method}
\subsection{Overview}
In the standard visual RL framework, we introduce VLM-based reasoning to enhance policy learning. Specifically, the VLM processes the current visual observation and task-specific prompts to infer action semantics, which are then mapped to continuous action $a_{v,t}$ via a string-to-float parsing function. We denote the native action of $\pi_{\theta}(\cdot)$ as $a_{r,t}$. Notably, only $a_{r,t}$ interacts with the environment during training, while $a_{v,t}$ serves as an auxiliary supervisory signal for policy refinement. During testing, our method relies exclusively on the actor network of the visual RL system for decision-making, ensuring that real-time performance requirements are met.

\begin{figure*}[!ht] 
  \centering
   \includegraphics[width=1.0\linewidth]{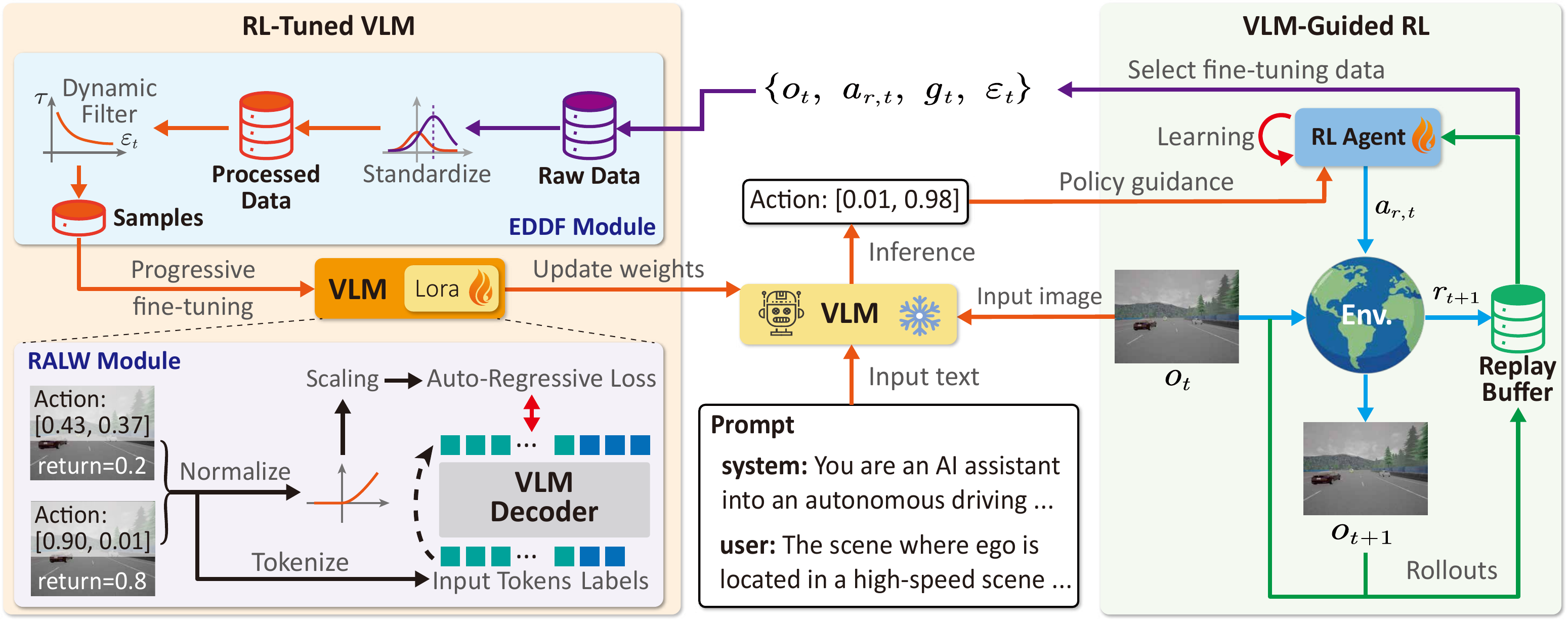}
   \caption{Collaborative optimization framework of \ours. It consists of two main components: (1) VLM-Guided RL. During this stage, the agent learns the policy under the guidance of actions inferred by the VLM. (2) RL-tuned VLM. This part comprises two essential modules: EDDF and RALW. Through the interaction of these two modules, the expertise of the VLM in specific domains is improved, which in turn benefits RL. }
   \label{fig:method}
\end{figure*}

The framework of \ours\ is shown in \Cref{fig:method}. \ours\ consists of two main components: (1) VLM-Guided RL. We adopt the standard SAC training procedure and use $a_{v,t}$ as a regularization constraint on the learning of $\pi_{\theta}(\cdot)$, enabling improved exploration and mitigation of suboptimal convergence. (2) RL-tuned VLM. In this component, the VLM is iteratively refined using RL-generated trajectories to enhance their scene understanding. 

First, due to the instability of RL training, many noisy or low-quality samples are generated. Therefore, we develop an EDDF module, which dynamically adjusts the threshold for sample selection based on the exploration level in RL, enabling more effective fine-tuning of the VLM. 

Second, the stochastic nature of exploration often leads to inconsistent actions under visually similar observations. For example, choosing between ``accelerate forward'' and ``decelerate and turn right'' on a straight road. Training the VLM directly on such noisy and inconsistent data may mislead supervised fine-tuning. To mitigate this, we introduce a RALW module, which assigns trajectory-specific loss weights proportional to their returns, prioritizing high-return trajectories and mitigating the interference from inconsistent actions under similar observations.

\subsection{Policy Guidance}
In VLM-Guided RL, the iterative improved the VLM provides policy-level guidance to the RL agent by leveraging its generalization capability and better domain-specific knowledge, leading to a more globally optimal policy. To implement this guidance, we follow the previous work~\cite{zhou2024large}, which introduces a regularization term on $\mathcal{L}_{\pi}$ that aligns the actions $a_{v,t}$ predicted by the VLM with those from $\pi_{\theta}(\cdot)$. Formally, the final loss of $\pi_{\theta}(\cdot)$ is defined as:
\begin{equation}
\mathcal{L}_{\tilde{\pi}}= \mathcal{L}_{\pi} + \lambda \left\|a_{v,t}-a_{r,t}\right\|_{2}^{2}, 
\label{eq:4}
\end{equation}
where $\lambda$ is the weight. 

\subsection{Exploration-Driven Dynamic Filter}\label{EDDF}
To identify high-quality samples for fine-tuning the VLM, we design the Exploration-Driven Dynamic Filtering (EDDF) module within the collaborative optimization framework. Given that immediate rewards $r_{t}$ reflect short-term gains and $Q$-value estimates may be biased, we adopt the trajectory-based cumulative return as the core metric for policy evaluation. Formally, the return is defined as $g_{t}$=$\sum_{k=t}^{T} \gamma^{k-t} r_{k}$, where $T$ is the terminal step of the trajectory. $g_{t}$ provides a more comprehensive assessment of policy performance. Specifically, EDDF operates in three stages: 

(1) \textbf{Data Storage}. We maintain a dedicated buffer $\mathcal{D}_{f}$ to store trajectory data for fine-tuning, including observations $o_{t}$, actions $a_{r,t}$, and returns $g_{t}$. Formally, $\mathcal{D}_{f}$=$\left\{\left(o_{i}, a_{r, i}, g_{i}\right)\right\}_{i=1}^{N}$, where $N$ is the length of $\mathcal{D}_{f}$.

(2) \textbf{Data Transformation}. Inspired by data normalization techniques in machine learning, we apply a Z-score transformation~\cite{cheadle2003analysis} to the return values in $\mathcal{D}_{f}$: $\mathcal{G}_{z}$=$\text{Z-score}(\left\{g_{i}\right\}_{i=1}^{N})$. This standardizes the distribution, enhances sensitivity to outliers, and ensures robustness across training phases. 

(3) \textbf{Dynamic Filtering}. The core idea is to adaptively adjust the filtering threshold $\tau$ based on the agent's exploration-exploitation behavior. During early training, when policy entropy $\varepsilon_{t}$ of visual RL is high, we retain more potentially valuable samples with a lower threshold, even if they yield low returns due to stochasticity. As training progresses and entropy decreases, the threshold becomes stricter to prioritize high-return trajectories as $\varepsilon_{t}$ decreases and tends to stabilize. The threshold $\tau$ is computed as:
\begin{equation}
\tau=\mathrm{Median}(\mathcal{G}_{z})+ \mathrm{Sigmoid}(\varepsilon_{t}) \cdot\mathrm{IQR}(\mathcal{G}_{z}),
\label{eq:5}
\end{equation}
where $\mathrm{Median}(\cdot)$ computes the median value of $\mathcal{G}_{z}$, $\mathrm{IQR}(\cdot)$ measures the interquartile range (25th–75th percentile spread)~\cite{vinutha2018detection}, and the Sigmoid function $\mathrm{Sigmoid}(\cdot)$ maps entropy $\varepsilon_{t}$ to a normalized scaling factor. This ensures $\tau$ dynamically responds to the agent's learning dynamics. When needing fine-tuning, we screen out the samples with $g_{i}$ exceeding the threshold $\tau$ in $\mathcal{D}_{f}$ and retrieve the final corresponding $\left(o_{i}, a_{r, i}, g_{i}\right)$ pairs based on the index of $g_{i}$ for fine-tuning the VLM. 

\begin{algorithm}[!ht]
\caption{Relevant training pipeline of \ours}
\label{Algorithm:1}
\begin{algorithmic}[1]
\STATE Initialize the SAC algorithm, the train steps $T_{train}$, \\ the vision-language model (VLM), corresponding prompt, Replay buffer $\mathcal{D}$, and $\mathcal{D}_{f}$;
\STATE Set $c$=0, $f_{t}$=0, and $\psi_{c+1}=\psi_{c} +\psi_{c}*c$;
\FOR{Every step $t$ in $T_{train}$}
    \IF{Game is done} 
        \STATE Reset environment;
    \ENDIF
    \STATE Execute the VLM to reason $a_{v,t}$ based on the \\ current observation image and prompt;
    \STATE Sample $a_{r,t} \sim \pi_\theta(\cdot)$ and execute it to get $r_{t+1}$, \\ $o_{t+1}$, and terminal state $d_t$;
    \STATE Store transition $(o_t, a_{r,t}, r_{t+1}, o_{t+1}, d_t, a_{v,t})$ in $\mathcal{D}$;
    \STATE Calculate $g_{t}$ and store $(o_t, a_{r,t}, g_{t})$ in $\mathcal{D}_{f}$;
    \IF{Enough samples in $\mathcal{D}$}
        \STATE Sample batch data from $\mathcal{D}$ to Update SAC.
    \ENDIF
    \IF{$f_{t}$ \% $\psi_{c+1}$ is 0}
        \STATE Select data from $\mathcal{D}_{f}$ using the EDDF module \\ to fine-tuning the VLM with the RALW module;
        \STATE Reset $f_{t}$=0 and clear $\mathcal{D}_{f}$;
        \STATE $c \gets c+1$\;
        \STATE $\psi_{c+1} \gets \psi_{c} +\psi_{c}*c$;
    \ENDIF
    \STATE $f_{t} \gets f_{t}+1$\;
    
\ENDFOR
\end{algorithmic}
\end{algorithm}

\subsection{Return-Aware Adaptive Loss Weight}
To address the issue of action inconsistency, we propose the Return-Aware Adaptive Loss Weighting (RALW) module that dynamically adjusts the influence of each sample during fine-tuning based on its cumulative reward. We first select high-quality samples via the EDDF module and normalize returns in the selected samples to the range $[-1, 1]$. Samples with negative returns ($g_{t} <$ 0) are assigned zero weight to suppress their influence, while high-return samples ($g_{t} >$ 0) receive higher weights to emphasize favorable behaviors. This design preserves the pre-trained VLM's basic capabilities while enabling performance improvement through high return-guided fine-tuning. 

Next, we construct a training batch $K$=$\left\{\left(x_{b}, y_{b}, \bar{g}_{b}\right)\right\}_{b=1}^{B}$, where $x_{b}$ is the input tokens (tokenized from the input prompt and $o_{t}$), $y_{b}$ is the label tokens (tokenized from $a_{r, t}$), $\bar{g}_b \in [-1,1]$ is the normalized return, and $B$ is the batch size. We incorporate external return signals into the fine-tuning process by formulating the objective as a return-weighted auto-regressive loss. Let $T$ be the token length, and $N_{\text{v}}$ represent a total number of valid token in the batch. Formally, the return-weighted auto-regressive loss can be defined as:
\begin{equation}
\mathcal{L}_{\mathrm{RALW}}=\frac{1}{N_{\text {v}}} \sum_{b=1}^{B} w_{b} \sum_{t=1}^{T}-\log p\left(y_{b, t} \mid \mathbf{x}_{b,<t}\right),
\label{eq:6}
\end{equation}
where $\log p\left(y_{b, t} \mid x_{b,<t}\right)$ refers to the model forecast $y_{b, t}$ of conditional probability, and $w_{b}$=$\max(\bar{g}_{b},0)$ denotes the non-negative weight. 
This formula ensures that the model prioritizes learning label tokens with higher returns, thereby reducing the interference that action bias may introduce under similar observations. Notably, the auto-regressive loss utilized is the negative log-likelihood (NLL) loss with label smoothing~\cite{guo2024cross}, where a label smoothing regularization will be incorporated into the auto-regressive loss to alleviate overfitting. 

\subsection{Progressive Fine-tuning}
To improve training efficiency and reduce computational overhead, we adopt a progressive fine-tuning strategy, where both the fine-tuning interval and the size of $\mathcal{G}_{z}$ gradually increase during training. This is motivated by the observation that frequent updates are necessary in the early stages when the policy is unstable, while less frequent updates suffice as the policy gradually converges. Specifically, we define a step size $\psi_{c}$ that increases linearly with the number of fine-tuning iterations $c$, that is, $\psi_{c+1}=\psi_{c} +\psi_{c}*c$, where $c$=$0,1,2,...$. Notably, $\mathcal{D}_{f}$ is cleared after each fine-tuning and a fine-tuning update is triggered only when the accumulated training steps reach $\psi$. In addition, we utilize Lora fine-tuning technology~\cite{hu2022lora} to further minimize the resource consumption. For clarity, \Cref{Algorithm:1} provides the relevant training pipeline of \ours.

\section{Experiments}

\subsection{Experimental Settings}
\paragraph{Environments.}
To evaluate the effectiveness of \ours\ in complex visual  tasks, we conducted tests on two widely used RL benchmarks: (1) the CARLA simulator~\cite{dosovitskiy2017carla} for autonomous driving, and (2) the DMControl~\cite{tassa2018deepmind} for robotic control. The CARLA simulator offers highly realistic visual inputs, which include a wide range of task-irrelevant details. This makes it an ideal benchmark for assessing algorithm performance in complex and realistic driving scenarios. For CARLA, we evaluate \ours\ on two challenging scenarios: (1) Highway (\#HW), with up to 10 random vehicles placed in front of the agent, and (2) Ghost Pedestrian (\#GP), where pedestrians suddenly cross the road behind static vehicles at three predefined locations. The agent must drive as far as possible within 1,000 time steps while avoiding collisions. For DMControl, we report experimental results on six commonly used tasks and other hard tasks, demonstrating the strong capability of \ours\ across different domains.

\paragraph{Implementation Details.}
For the VLM, we employed Qwen2.5-VL-3B~\cite{bai2025qwen2} for prior knowledge reasoning tasks in the baseline experiments. 
Following the prior work~\cite{xu2025vlms}, experiments are conducted using three seeds and the mean and standard deviation of metrics are reported. For the parameters, we set $\lambda$=2.0 and initial $\psi_{0}$=5000 in the experiments. We trained each benchmark in 100K steps and reported the metrics over 10 evaluated episodes with different seeds. Each episode has a maximum of 1000 steps. The metrics in CARLA are the episode reward (ER) and the driving distance (DD) of the evaluated episodes. More implementation details are shown in the Appendix. 

\begin{table}[!t]
\normalsize 
\centering
\setlength{\tabcolsep}{4pt}
\begin{tabular}{c|c|cc}
\toprule 
\multirow{1}{*}{Type} & \multirow{1}{*}{Methods} & ER $\uparrow$  & DD $\uparrow$ \\
\midrule
\multirow{9}{*}{Vanilla visual RL}
& SAC & 69 ± 46	& 91 ± 56 \\
& DeepMDP & 155 ± 84 & 167 ± 89 \\
& CURL & 135 ± 63 & 152 ± 70\\
& DrQ & 115 ± 51 & 127 ± 55\\
& SPR & 84 ± 53 & 100 ± 61\\
& MLR & 106 ± 69 & 130 ± 80\\
& PER & 159 ± 68 & 175 ± 73 \\
& ERE & 117 ± 89 & 132 ± 95\\
& ResAct  & 227 ± 36 &  236 ± 40 \\
\midrule
\multirow{1}{*}{Only the VLM}
& VBE & -11 ± 5  & 11 ± 4 \\
\midrule
\multirow{5}{*}{\makecell{VLM-assisted \\ visual RL}} 
& DPL  & 113 ± 63 &	124 ± 67 \\
& APL  & 146 ± 74	& 155 ± 76 \\
& VPF  & 91 ± 24	&  99 ± 24  \\
& DGC  & 208 ± 13 & 234 ± 15 \\
& \ours\ (Ours) & \cellcolor{gray!20}\textbf{248 ± 81} & \cellcolor{gray!20}\textbf{259 ± 85} \\
\bottomrule
\end{tabular}
\caption{Performance comparison in the \#HW scenario. The best results for each metric are denoted by \raisebox{0.5ex}
{\colorbox{gray!20}{\;}}.}
\label{tab:result_on_carla_1}
\end{table}

\subsection{Comparison with State-of-the-art}
\paragraph{Results on CARLA.} 
First, we evaluate \ours\ against various visual RL methods, including SAC~\cite{haarnoja2018soft2}, auxiliary loss-based methods (CURL~\cite{laskin2020curl}, MLR~\cite{yu2022mask}), a data augmentation method (DrQ~\cite{yarats2021image}), motion modeling methods (DeepMDP~\cite{gelada2019deepmdp}, SPR~\cite{schwarzer2020data}, ResAct~\cite{liu2025visual}), and buffer-based optimization methods (PER~\cite{schaul2015prioritized}, ERE~\cite{wang2019boosting}). In addition, we compare the VLM-based methods in the same direction, including a VLM-based executor (VBE~\cite{mei2024replanvlm}), the methods of directly adding prior loss (DPL~\cite{xu2024vlm}, DGC~\cite{xu2025vlms}), a method of annealing weight-based policy loss (APL~\cite{zhou2023large, lee2025sample}), and a VLM-based policy fine-tuning method (VPF~\cite{zhai2024fine, wei2025gtr}). 

As presented in \Cref{tab:result_on_carla_1} and \Cref{tab:result_on_carla_2}, \ours\ improves on both the episode reward and driving distance. We make the following conclusions: (1) VBE underperforms due to the inherent limitations of the basic VLM in adapting to continuous and dynamic environments; (2) The methods, including DPL, APL, and DGC, exhibit limited performance due to suboptimal knowledge transfer from unrefined VLMs; (3) Vanilla visual RL methods struggle to adapt to complex scenarios and thus have low performance; (4) \ours\ provides more reliable policy guidance through the collaborative optimization framework, enabling smoother and more stable vehicle control.

\begin{table}[!t]
\normalsize
\centering
\setlength{\tabcolsep}{4pt}
\begin{tabular}{c|c|cc}
\toprule 
\multirow{1}{*}{Type} & \multirow{1}{*}{Methods} & ER $\uparrow$  & DD $\uparrow$  \\
\midrule
\multirow{9}{*}{Vanilla visual RL}
& SAC & 38 ± 29 & 40 ± 29 \\
& DeepMDP & 82 ± 51 & 85 ± 52 \\
& CURL & 78 ± 52  & 81 ± 53\\
& DrQ & 110 ± 68 & 112 ± 69\\
& SPR & 62 ± 42 &  66 ± 45\\
& MLR & 118 ± 50 & 121± 51\\
& PER & 51 ± 45 & 54 ± 46\\
& ERE & 55 ± 50 & 58 ± 53\\
& ResAct & 212 ± 54 & 216 ± 55 \\
\midrule
\multirow{1}{*}{Only the VLM}
& VBE & 9 ± 7	& 18 ± 7 \\
\midrule
\multirow{5}{*}{\makecell{VLM-assisted \\ visual RL}} 
& DPL  & 127 ± 51 & 129 ± 52 \\
& APL  & 107 ± 54	& 111 ± 55 \\
& VPF  & 81 ± 2 &  82 ±  3 \\
& DGC  & 146 ± 14 &  169 ± 18 \\
& \ours\ (Ours) & \cellcolor{gray!20}\textbf{235 ± 89} & \cellcolor{gray!20}\textbf{237 ± 89}  \\
\bottomrule
\end{tabular}
\caption{Performance comparison in the \#GP scenario. The best results for each metric are denoted by \raisebox{0.5ex}
{\colorbox{gray!20}{\;}}.}
\label{tab:result_on_carla_2}
\end{table}

\paragraph{Results on DMControl.}
For DMControl, we first introduce \ours\ on several typical different baselines including SAC~\cite{haarnoja2018soft2}, 
DeepMDP~\cite{gelada2019deepmdp}, and RAD~\cite{laskin2020reinforcement}. As presented in \Cref{tab:result_on_dmc_1}, \ours\ consistently improves the performance of multiple baseline methods, highlighting its strong adaptability. 

We select RAD+\ours\ as our baseline. Then, we select a range of SOTA methods for comparison with \ours\, including (1) model-free methods: CURL, DrQ, SVEA~\cite{hansen2021stabilizing}, PlayVirtual~\cite{yu2021playvirtual}, TACO~\cite{zheng2023texttt}, MLR, MADI~\cite{grooten2024madi}, PSRL~\cite{choi2023local}, ResAct, and (2) model-based methods: Dreamer~\cite{hafner2019dream}, PlaNet~\cite{hafner2019learning}. We also present the results of previous VLM-based methods, including VBE, DPL, APL, and VPF. As reported in \Cref{tab:result_on_dmc_2}, \ours\ achieves new state-of-the-art. Furthermore, the significantly lower reward standard deviation indicates that \ours\ also offers improved convergence stability and optimization ease. The results of hard tasks can be found in the Appendix.

\begin{table*}[!ht]
\setlength{\tabcolsep}{1mm}
\centering
\setlength{\tabcolsep}{4pt}
\begin{tabular}{c|cccccc}
\toprule 
\multirow{1}{*}{Tasks/Methods} & SAC  & SAC+\ours\ &  DeepMDP & DeepMDP+\ours\ & RAD & RAD+\ours\ \\
\midrule  
Cartpole, Swingup & 237 ± 49  & \cellcolor{gray!20}\textbf{740 ±108} & 389 ± 44  & \cellcolor{gray!20}\textbf{793 ± 26} & 694 ± 28  & \cellcolor{gray!20}\textbf{872 ± 2} \\
Reacher, Easy & 239 ± 183  & \cellcolor{gray!20}\textbf{246 ± 345} & 471 ± 173  & \cellcolor{gray!20}\textbf{461 ± 443} & 734 ± 87  & \cellcolor{gray!20}\textbf{969 ± 18}\\
Cheetah, Run & 118 ± 13  & \cellcolor{gray!20}\textbf{156 ± 29} & 306 ± 25  & \cellcolor{gray!20}\textbf{352 ± 35} & 364 ± 38  & \cellcolor{gray!20}\textbf{504 ± 13}\\
Walker, Walk & 95 ± 19  & \cellcolor{gray!20}\textbf{194 ± 57} & 384 ± 197  & \cellcolor{gray!20}\textbf{397 ± 71} & 552 ± 87  & \cellcolor{gray!20}\textbf{802 ± 25}\\
Finger, Spin & 230 ± 194  & \cellcolor{gray!20}\textbf{247 ± 51}
 & 509 ± 72  & \cellcolor{gray!20}\textbf{653 ± 18}
 & 813 ± 65  & \cellcolor{gray!20}\textbf{976 ± 9}\\
Ball in cup, Catch & 85 ± 130  & \cellcolor{gray!20}\textbf{185 ± 331} & 704 ± 24 & \cellcolor{gray!20}\textbf{757 ± 169} & 825 ± 49 & \cellcolor{gray!20}\textbf{960 ± 23}
 \\
\bottomrule
\end{tabular}
\caption{Implementation of \ours\ on top of three different baselines. +\ours\ represents the preceding method combined with \ours.}
\label{tab:result_on_dmc_1}
\end{table*}

\begin{table*}[!ht]
\small
\centering
\setlength{\tabcolsep}{4pt}
\begin{tabular}{c|c|cccccc}
\toprule 
\multirow{1}{*}{Type} & \multirow{1}{*}{Methods} & Cartpole, Swingup  & Reacher, Easy &  Cheetah, Run & Walker, Walk & Finger, Spin & Ball in cup, Catch \\
\midrule
\multirow{11}{*}{Vanilla visual RL}
& CURL & 582 ± 146&	538 ± 233&	299 ± 48&	403 ± 24&	767 ± 56 & 769 ± 43 \\
& DrQ & 759 ± 92&	601 ± 213&	344 ± 67&	612 ± 164&	901±104 & 913±53\\
& SVEA & 727 ± 86&	811 ± 115&	375 ± 54&	747 ± 65&	859 ± 77 & 915 ± 71\\
& PlayVirtual & 816 ± 36 & 785 ± 142 & 474 ± 50	& 460 ± 173 &915 ± 49 & 929 ± 31\\
& TACO & 782 ± 51& 821 ± 97& 402 ± 62& 601 ± 103& 876 ± 67  & 902 ± 54\\
& MLR & 806 ± 48&	866 ± 103&	482 ± 38&	643 ± 114&	907 ± 58  & 933 ± 16\\
& MADI & 704 ± 54& 766 ± 101& 432 ± 44& 574 ± 94& 810 ± 95  & 884 ± 36\\
& PSRL & 849 ± 63& 621 ± 202& 398 ± 71& 595 ± 104& 882 ± 132  & 922 ± 60\\
& ResAct &819 ± 44& 917 ± 59& 503 ± 42& 772 ± 65& 974 ± 42  & 948 ± 44\\
& Dreamer & 326±27& 314±155&	235±137&	277±12&	341±70  & 246±174\\
& PlaNet & 563±73&	82±174&	 165±123&	224±48	& 560±77 & 0±0 \\
\midrule
\multirow{1}{*}{Only the VLM}
& VBE & 178 ± 31&	94  ± 138&	3 ± 2&	26 ± 11& 0 ± 0 & 49  ± 213 \\
\midrule
\multirow{4}{*}{\makecell{VLM-assisted \\ visual RL}} 
& DPL  & 776 ± 5 & 224 ± 387 &	255 ± 16 &	209 ± 64 &	815 ± 8  & 751 ± 148  \\
& APL  & 820 ± 3 & 292 ± 436 & 367 ± 30	& 696 ± 48	& 839 ± 9  & 905 ± 95 \\
& VPF  & 749 ± 21  & 162 ± 137 & 127 ± 2  & 132 ± 4 & 0 ± 0  & 0 ± 0  \\
& \ours\ (Ours) & \cellcolor{gray!20}\textbf{872 ± 2} & \cellcolor{gray!20}\textbf{969 ± 18}  & \cellcolor{gray!20}\textbf{504 ± 13
} & \cellcolor{gray!20}\textbf{802 ± 25}  & \cellcolor{gray!20}\textbf{976 ± 9} & \cellcolor{gray!20}\textbf{960 ± 23}\\
\bottomrule
\end{tabular}
\caption{Performance comparison with SOTA methods in DMControl. The best results for each metric are denoted by \raisebox{0.5ex}
{\colorbox{gray!20}{\;}}.}
\label{tab:result_on_dmc_2}
\end{table*}			

\begin{figure}[!t]
\centering
\begin{subfigure}{0.495\linewidth}
    \includegraphics[width=\linewidth]{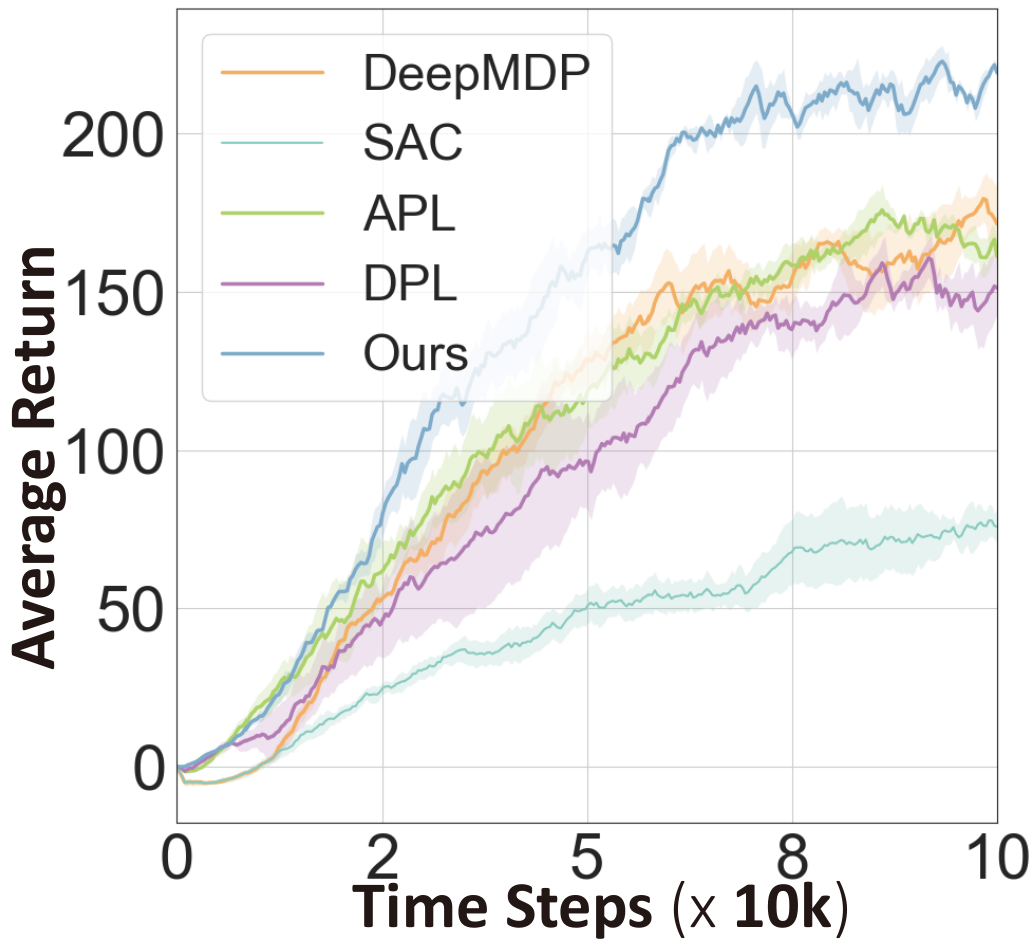}
    \caption{The \#HW scenario.}
\end{subfigure}
\begin{subfigure}{0.495\linewidth}
    \includegraphics[width=\linewidth]{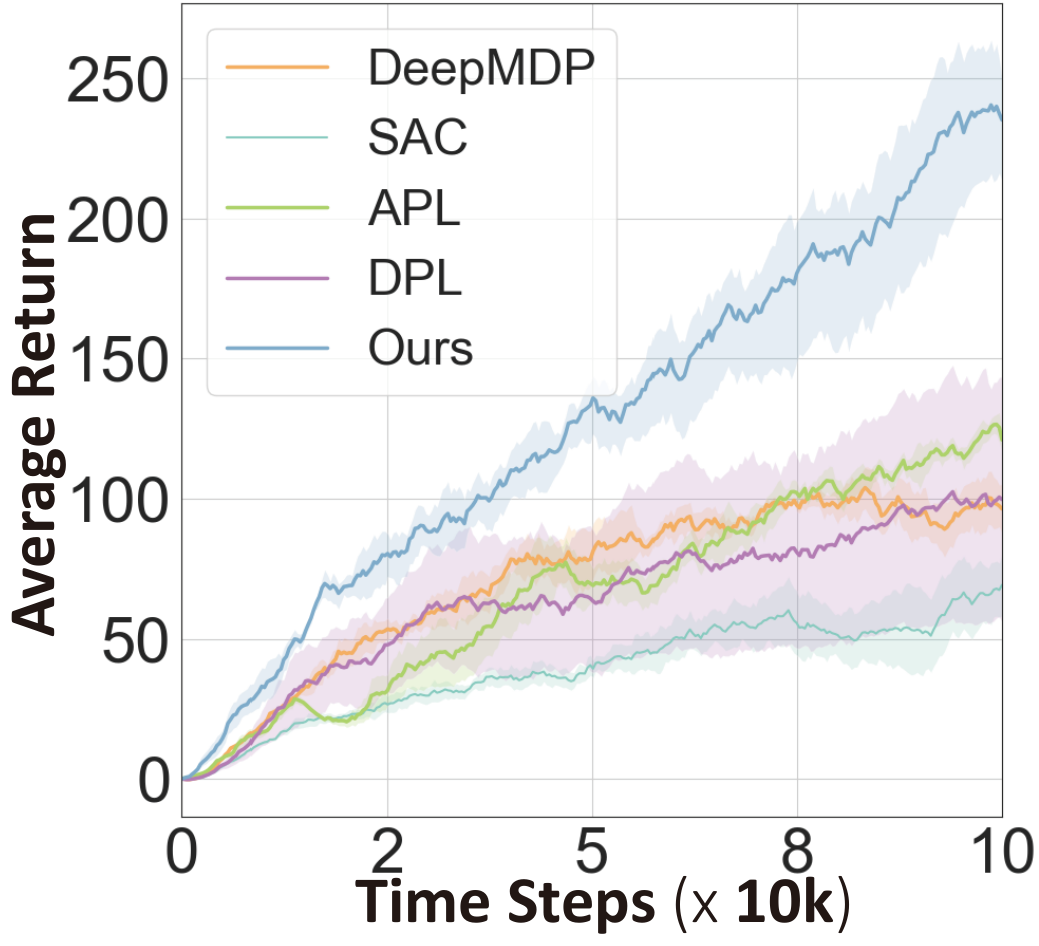}
    \caption{The \#GP scenario.}
\end{subfigure}
\caption{Visualization of the average return curve during training in CARLA.}
\label{fig:vis_my}
\end{figure}

\subsection{Ablation Study}
To assess the contributions of each component of \ours, we conduct ablation studies for the \#HW scenario. The results are summarized in \Cref{tab_ablation}. We can draw the following analyses: 

\paragraph{Effects of EDDF module.}
The results of \textbf{M1} demonstrate that the EDDF module outperforms random data filtering by preventing the VLM from learning suboptimal knowledge. \textbf{M2}–\textbf{M4} further validate EDDF’s superiority over fixed top-k methods: rigid thresholding risks missing high-quality actions or introducing noisy samples. In contrast, EDDF dynamically adapts thresholds to capture low-return yet valuable actions during exploration, thereby improving the knowledge density and generalization of the VLM. \textbf{M5} shows that Z-score normalization enhances data stability and contributes to more robust learning. \textbf{M6} shows that evaluating performance based on trajectory-level cumulative returns yields better insights than per-step immediate rewards. \textbf{M7} indicates that, due to the stochastic nature of RL exploration, unstable initial $Q$-value estimation may lead to the selection of low-quality samples, which in turn degrades the fine-tuning performance of the VLM and negatively impacts policy learning. 

\paragraph{Effects of RALW module.} 
Experiments on \textbf{M8} and \textbf{M9} validate the proposed RALW module. The results indicate that its performance cannot be replicated using randomly generated returns, highlighting the necessity of learning adaptive loss weights based on the dynamic returns. 

\paragraph{Effects of the VLM.} 
To highlight the importance of VLM guidance, we design a baseline in \textbf{M10} and \textbf{M11} that selects top-return samples (80\% and 20\% per batch) for visual reinforcement learning training, replacing the rest with random actions to balance exploration and exploitation. Despite this strategy, it underperforms \ours, confirming the critical role of enhancing VLM task knowledge and leveraging its generalization for policy guidance.

In the Appendix, we further evaluate various experiments on \ours. These experiments offer additional evidence of \ours’s effectiveness.

\begin{table*}[!ht]
\small
\centering
\setlength{\tabcolsep}{4pt}
\begin{tabular}{c|c|cc|l}
\toprule
    Type &
    \multirow{1}{*}{\textbf{Idx.}} & 
    \multirow{1}{*}{\makecell[c]{ER $\uparrow$}} & 
    \multirow{1}{*}{\makecell[c]{DD $\uparrow$}} & 
    \multirow{1}{*}{\makecell[c]{Description}} \\

\midrule
\multirow{8}{*}{Effects of EDDF module.}  
  & \textbf{M1} & 144 ± 74 & 155 ± 77 & W/o EDDF module, i.e., randomly filtering data for fine-tuning. \\
  &  \textbf{M2} & 204 ± 137 & 214 ± 141 & Replace the EDDF module with the return-based top-80\% method. \\
  &  \textbf{M3} & 217 ± 95 & 229 ± 98 & Replace the EDDF module with the return-based top-90\% method. \\
  &  \textbf{M4} & 192 ± 107 & 203 ± 110 & Replace the EDDF module with the return-based top-95\% method. \\
  &  \textbf{M5} & 210 ± 99 & 221 ± 102 & W/o Z-score to standardize the data. \\
  &  \textbf{M6} & 221 ± 105 & 231 ± 108 & Change return to reward. \\
  &  \textbf{M7} & 200 ± 118 & 212 ± 122 & Change return to $Q$-value .calculated by $Q_{\phi}(\cdot)$. \\
  
\midrule
\multirow{2}{*}{Effects of RALW module.} 
  &  \textbf{M8}  & 204 ± 111 & 214 ± 115 & W/o RALW module. \\
  &  \textbf{M9}  & 184 ± 78 & 191 ± 81 & The loss weight is randomly generated instead of return-based guidance. \\

\midrule
\multirow{2}{*}{Effects of the VLM.} 
  &  \textbf{M10} & 183 ± 96 & 195 ± 99 & Select top-return (80\%) and randomly mixed samples to train RL. \\
  &  \textbf{M11} & 175 ± 58 & 185 ± 58 & Select top-return (50\%) and randomly mixed samples to train RL. \\
\midrule
 \cellcolor{gray!20} Ours &  \cellcolor{gray!20} \ours &  \cellcolor{gray!20}\textbf{248 ± 81} & \cellcolor{gray!20}\textbf{259 ± 85} & \cellcolor{gray!20} Full version.\\
\bottomrule
\end{tabular}
\caption{Ablation experiments conducted in the \#HW scenario validate the effectiveness of \ours. The best results for each metric are denoted by \raisebox{0.5ex}{\colorbox{gray!20}{\;}}.}
\label{tab_ablation}
\end{table*}

\subsection{Visual Analysis}
\Cref{fig:vis_my} presents the average return curve during training of some SOTA methods in the CARLA scenarios. As shown, \ours\ effectively leverages the powerful reasoning and generalization capabilities of the VLM, significantly enhancing policy learning efficiency in visual RL. 
In contrast, other baseline methods fail to achieve superior global performance due to the lack of effective self-exploration mechanisms or limitations imposed by their VLMs' guidance.
Additional visual analyses in the Appendix, such as VLM action reasoning differences before/after fine-tuning and quantitative evaluation of the EDDF module, further validate the effectiveness of \ours. 

\section{Conclusion}
We have proposed a novel LM-assisted RL method that features a collaborative optimization paradigm. It comprises two key modules: one for effective sample selection from RL trajectories, and another for mitigating action inconsistency during fine-tuning. In addition, we have introduced a progressive fine-tuning approach to reduce resource consumption and enhance efficiency. Experiments across various complex and high-dimensional environments have shown that \ours\ achieves state-of-the-art performance. 

\section*{Acknowledgments}
The study was funded by the Shenzhen Basic Research Fund under grant JCYJ20241202130025030; Shenzhen Science and Technology Program (KQTD20240729102051063); the National Natural Science Foundation of China under contracts No. 62422602, No. 62372010, No. 62425101, No. 62332002, No. 62206281; Key Laboratory Grants 241-HF-D05-01; and the major key project of the Peng Cheng Laboratory (PCL2021A13 and PCL2025A02). Computing support was provided by
Pengcheng Cloudbrain.

\bibliography{aaai2026}

\clearpage
\appendix 
\section*{Appendix}

\subsection{More Explanations of Motivation}
\begin{figure}[!ht] 
  \centering
   \includegraphics[width=1.0\linewidth]{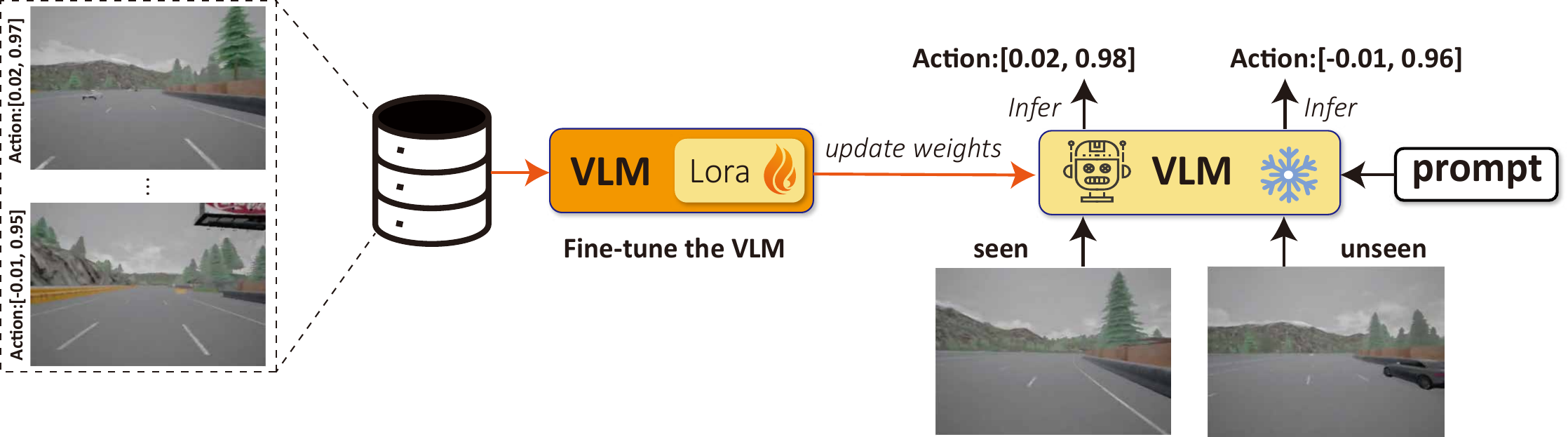}
   \caption{Schematic illustration of VLMs' generalization capability. After fine-tuning, the model demonstrates the ability to generalize its inference performance to unseen yet structurally similar scenarios, such as both being straight road conditions.}
   \label{fig:mo}
\end{figure}

The core motivation of our approach lies in harnessing the multimodal understanding and strong generalization capabilities of VLMs to guide visual RL. As illustrated in \Cref{fig:mo}, we fine-tune the VLM using high-quality image-action pairs selected via our designed EDDF module and employing Lora for parameter-efficient adaptation. To evaluate generalization, we perform inference on both seen (training time) and unseen (novel) images. On a representative driving scenario, the model outputs [0.02,0.98] and [-0.01,0.96] for steer and throttle, respectively. Despite the unseen observation, the action remains coherent and close to the seen case, demonstrating the VLM’s ability to generalize across visual variations.

This property enables \ours\ to transfer semantic and behavioral priors from VLMs into the RL process more efficiently. While RL trajectories may still contain suboptimal transitions, they are generally more reliable than the VLM’s initial inference, especially in dynamic environments. By establishing a bidirectional optimization loop between VLM and RL, \ours\ introduces a principled framework for collaborative visual decision-making, advancing beyond unidirectional VLM-to-RL knowledge transfer.

\subsection{More Detailed Implementation}\label{CARLA_benchmarks}
\subsubsection{Basic Visual RL Algorithm.}
We focus on complex visual continuous control tasks. In DMControl, our method is an improvement based on the RAD method~\cite{laskin2020reinforcement}. While in the CARLA environment, our method builds upon SAC~\cite{haarnoja2018soft1, haarnoja2018soft2}. \ours\ enhances SAC by incorporating a transition network from DeepMDP~\cite{gelada2019deepmdp}, which predicts both the next state and reward, thereby promoting more effective visual encoder learning through model-based supervision.

\subsubsection{CARLA Benchmarks.}
The agent’s action space in CARLA is defined as [$steer$, $throttle$]. The steering control ($steer$) ranges from -1 (left turn) to 1 (right turn), with 0 indicating straight driving. Throttle ($throttle$) also lies in [-1, 1], where values $\le$ 0 correspond to braking or deceleration, and > 0 indicate acceleration.

In the visual RL setup, the agent receives RGB images as input. To ensure consistency across experiments, we fixed the camera parameters under uniform weather conditions (Cloudy Noon). Some key hyperparameters of the RGB camera settings and experiments in the CARLA benchmarks are provided in \Cref{tab:rgb} and \Cref{tab:hyper}. It is worth noting that, to minimize inference latency, the VLMs are invoked once every 10 frames during execution.

\begin{table*}[!ht]
\normalsize
\centering
\begin{tabular}{l|ll}
\toprule
    Attributes & Value & Description \\
\midrule
    image size & [128, 128] & Width and height of the image in pixels. \\
    fov & 60 & Horizontal field of view (FOV) of the camera. \\
    tick & 20 & The RGB camera's capture frequency in hertz. \\
    gamma & 2.2 & The gamma correction applied to the RGB camera's output. \\
    iso & 100 & The camera sensor sensitivity. \\
    exposure\_min\_bright & 10 & Minimum brightness for auto exposure.  \\
    exposure\_max\_bright & 12 & Maximum brightness for auto exposure.  \\
    motion\_blur\_intensity & 0.45 & Strength of motion blur. \\
\bottomrule
\end{tabular}
\caption{Key parameters of the RGB camera in the CARLA benchmarks.}
\label{tab:rgb}
\end{table*}

\begin{table*}[!t]
\normalsize
\centering
\begin{tabular}{l|l}
\toprule
    Hyperparameter & Value \\
\midrule
    RGB frame dimensions & 128$\times$128$\times$3  \\
    Action repeat & 4 \\
    Frame stack & 3 \\
    Initial sampling steps (warming up) & 1,000 \\
    Total training steps & 100,000 \\
    Evaluation episodes & 10 \\
    Replay buffer size & 100,000 \\
    Initial $\alpha$ ($\alpha_{0}$)  & 0.1 \\
    Learning rate of $\alpha$  & $10^{-4}$ \\
    Learning rate of $\pi_{\theta}$  & $10^{-3}$ \\
    Learning rate of $Q_{\phi}$  & $10^{-3}$ \\
    Optimizer for $\pi_{\theta}$, $Q_{\phi}$ and $\alpha$ & Adam (betas=(0.9, 0.999)) \\
    Batch size & 128 \\
    VLMs-guided supervision update frequency & 10 \\
    Transition network update frequency & 1 \\
    $\pi_{\theta}$ update frequency & 2 \\
    $Q_{\phi}$ target update frequency & 2 \\
\bottomrule
\end{tabular}
\caption{Some key hyperparameters used in the Carla benchmarks.}
\label{tab:hyper}
\end{table*}

Our goal is to maximize travel distance within 100K training steps while minimizing collisions. We designed distinct reward functions for the \#HW and \#GP scenarios:

(1) Highway (\#HW):
\begin{equation}
    r_t = \lambda_1\cdot v \cdot d_t
        - \lambda_2\cdot collision 
        - \lambda_3\cdot |steer|,
\end{equation}
where $v$ is velocity, $d_t$ is time difference, and $collision$ denotes collision intensity from CARLA’s sensor. Coefficients are set as $\lambda_1=1$, $\lambda_2=10^{-4}$, and $\lambda_3=1$. This encourages fast and stable driving while penalizing collisions and excessive steering.

(2) Junctions (\#GP):
\begin{equation}
    r_t = \lambda_1\cdot v \cdot d_t
        - \lambda_2\cdot collision 
        - \lambda_3\cdot lane\_invasion
        - \lambda_4\cdot |steer|,
\end{equation}
This extends the highway reward by adding a penalty for lane invasion, detected via CARLA’s built-in lane sensor. Coefficients are $\lambda_1=1$, $\lambda_2=10^{-3}$, $\lambda_3=10^{-2}$, and $\lambda_4=0.1$, emphasizing safer navigation in complex junctions.

\begin{figure}[!ht]
\centering
\begin{subfigure}{0.45\linewidth}
    \includegraphics[width=\linewidth]{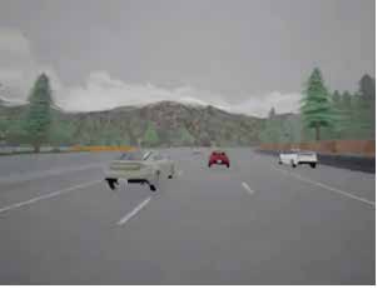}
    \caption{The \#HW scenario.}
\end{subfigure}
\begin{subfigure}{0.45\linewidth}
    \includegraphics[width=\linewidth]{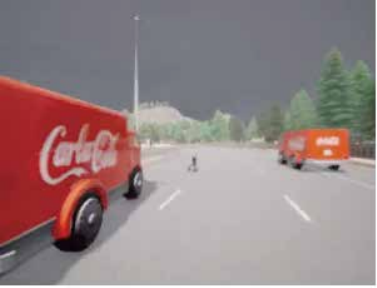}
    \caption{The \#GP scenario.}
\end{subfigure}
\caption{Visualization of the CARLA scenarios.}
\label{fig:vis_scenario}
\end{figure}

The visualizations of the two complex scenarios set by CARLA are shown in \Cref{fig:vis_scenario}. Moreover, the suboptimal quality of the early RL policy may generate low-quality samples, which may negatively impact the fine-tuning of VLMs. To mitigate this issue, we introduce a cold-start strategy with two optional components: 
(1) We delay the activation of VLMs-based policy guidance (as defined in \Cref{eq:4}) until after two rounds of fine-tuning in the \#HW. This preserves autonomous exploration during the early stages of visual RL and ensures that the VLMs provide more reliable guidance once sufficiently adapted. 
(2) Instead of using purely random actions for exploration during the warm-up phase, we utilize VLMs-generated reasoning to guide action selection, thereby improving the quality of early-stage interactions.

\subsubsection{DMControl Benchmarks.}
To ensure a fair comparison, we employed the same model architecture and hyperparameter configuration as RAD~\cite{laskin2020reinforcement}. We set $\lambda$=1.0 and initial $\psi_{0}$=5000 on six commonly used tasks. Similar to the settings of CARLA, we delay the activation of VLMs-based policy guidance in \Cref{eq:4} until after four rounds of fine-tuning. Some hyperparameters of the DMControl benchmarks are shown in \Cref{tab:DMControl}.

\begin{table*}[!ht]
\normalsize
\centering
\begin{tabular}{l|l}
\toprule
    \textbf{Hyperparameter} & \textbf{Value} \\
\midrule
Augmentation & Crop-walker, walk; Translate -otherwise \\
Observation rendering &  (100,100)  \\
Observation down/upsampling &  (84,84)(crop);(108,108)(translate)  \\
Replay buffer size & 100000 \\
Initial steps & 1000 \\
Stacked frames & 3 \\
Action repeat & 2 for finger, spin;  
8 for cartpole, swingup;
4 for otherwise  \\
Hidden units (MLP) & 1024 \\
Evaluation episodes & 10 \\
Optimizer & Adam \\
Beta $\left(\beta_{1}, \beta_{2}\right) \rightarrow\left(\pi_{\theta}, Q_{\phi}\right)$  & (0.9, 0.999) \\
Beta $ \left(\beta_{1}, \beta_{2}\right) \rightarrow(\alpha)$  & (0.5, 0.999) \\
Learning rate ($\pi_{\theta}$,$ Q_{\phi}$  ) &  2e-4  cheetah, run  1e-3  otherwise \\
Learning rate (  $\alpha$  ) &  1e-4  \\
 Batch Size & 512 \\
Critic target update freq & 2 \\
Convolutional layers & 4 \\
Number of filters & 32 \\
Non-linearity & ReLU \\
Discount  $\gamma$  & 0.99 \\
Initial temperature & 0.1 \\
\bottomrule
\end{tabular}
\caption{Some hyperparameters of the DMControl benchmarks.}
\label{tab:DMControl}
\end{table*}

\subsubsection{Fine-tuning of Lora Parameters.}
This work employs Lora-based fine-tuning to reduce the number of trainable parameters and, consequently, the overall training cost. Key Lora configuration details are summarized in \Cref{tab:Lora}. Given that VLMs possess a certain level of pre-existing knowledge relevant to autonomous driving, we set a relatively large $lora\_alpha$ to preserve as much of the original knowledge as possible during fine-tuning. In contrast, VLMs exhibit limited performance on DMControl tasks, where domain-specific knowledge is less transferable. Therefore, a smaller $lora\_alpha$ is used to allow the model to focus on learning new task-specific representations from scratch.  The number of fine-tuned parameters accounts for at most 8.2566\%. Other key parameters are empirical values that have been adjusted multiple times based on previous work~\cite{zhai2024fine}. At the same time, the selected parameters are all working and can be extended to other tasks.

\begin{table*}[!ht]
\normalsize
\centering
\begin{tabular}{l|l}
\toprule
    \textbf{Hyperparameter} & \textbf{Value} \\
\midrule
Target modules for CARLA & $down\_proj$, $o\_proj$, $q\_proj$, $up\_proj$, $k\_proj$, $gate\_proj$ \\
Target modules for others & $down\_proj$, $o\_proj$, $q\_proj$, $up\_proj$, $k\_proj$, $gate\_proj$, $v\_proj$, $qkv$, $lm\_head$ \\
$lora\_rank$ & 128  \\
$lora\_alpha$ & 256 for CARLA; 16 for others  \\
$lora\_dropout$ &  0.05 \\  
Learning rate (lr) &  1e-4 \\  
Batch size &  4 \\  
gradient accumulation steps &  4 \\  
number of train epochs & 2 \\  
type of lr scheduler  & cosine \\  
gradient checkpointing  & True \\  
\bottomrule
\end{tabular}
\caption{Key parameters of Lora-based fine-tuning.}
\label{tab:Lora}
\end{table*}

\subsubsection{Description of VLM-Assisted Visual RL Baselines.}
For a fair comparison, we introduce the previous method of enhancing visual RL through knowledge distillation based on LMs for comparison, which includes:

(1) VLM-Based Executor (VBE~\cite{mei2024replanvlm}). This method employs a vision-language model (VLM) to perform action inference and directly executes the inferred actions for decision-making without additional policy adaptation.

(2) Direct Policy Loss Integration (DPL~\cite{xu2024vlm}). DPL incorporates VLM-derived knowledge as a regularization term in the policy loss function of visual RL. The regularization weight $\lambda$ remains fixed throughout training. In our experiments, $\lambda$ was set equal to the baseline value used in \ours. This approach enables a principled comparison of how fine-tuned versus frozen VLMs serve as knowledge priors to guide policy learning in visual reinforcement learning. 

(3) Annealed Weight Policy Loss (APL~\cite{zhou2023large, lee2025sample}). Similar to DPL, APL integrates VLM knowledge as a regularization term but introduces a scheduled decay mechanism for $\lambda$. Specifically, we initialized $\lambda$=2 and reduced it by 0.01 every 200 training steps, with a minimum value of $\lambda$=0.1.

(4) VLM-Based Policy Fine-Tuning (VPF~\cite{zhai2024fine, wei2025gtr}). VPF treats the VLM as the policy network of RL and applies PPO~\cite{schulman2017proximal} for fine-tuning. This approach optimizes VLM-inferred actions through direct environmental interaction. Experimental settings followed the configurations in ~\cite{zhai2024fine}.

\subsection{Additional Experimental Results}

\subsubsection{Experiments on CarRacing.}
To further evaluate the generalization of our method across different autonomous driving environments, we conducted experiments on the CarRacing game in OpenAI Gym~\cite{brockman2016openai} based on pixel observations. Similar to CARLA, CarRacing is a driving control task, but with three continuous action dimensions: [steering, accelerator, brake], each ranging from -1 to 1. Unlike CARLA’s front-view perspective, CarRacing uses a top-down 2D view, which introduces a distinct observational setting and tests the robustness of \ours\ to varying visual inputs.

We trained the model with three random seeds under settings largely consistent with those used in CARLA. The results, summarized in \Cref{tab:result_on_carracing}, show that \ours\ still achieves the best performance, demonstrating its effectiveness across diverse environments. 

\begin{table}[!ht]
\normalsize
\centering
\setlength{\tabcolsep}{4pt}
\begin{tabular}{c|c|c}
\toprule 
\multirow{1}{*}{Type} & \multirow{1}{*}{Models} & ER $\uparrow$ \\
\midrule
\multirow{2}{*}{Vanilla visual RL}
& SAC & 287 ± 201 \\
& DeepMDP & 356 ± 169 \\
& CURL & 462 ± 201 \\
& DrQ & 354 ± 213 \\
& SPR & 380 ± 229 \\
& MLR & 350 ± 186 \\
\midrule
\multirow{1}{*}{Only VLMs}
& VBE & -18 ± 21 \\
\midrule
\multirow{4}{*}{\makecell{VLMs-assisted \\ visual RL}} 
& DPL  & 72 ± 84 \\
& APL  & 157 ± 149 \\
& VPF  & 12 ± 6 \\
& \ours\ (Ours) & \cellcolor{gray!20}\textbf{619 ± 221} \\
\bottomrule
\end{tabular}
\caption{Testing performance comparison with SOTA methods in CarRacing. The best results for each metric are denoted by \raisebox{0.5ex}
{\colorbox{gray!20}{\;}}.}
\label{tab:result_on_carracing}
\end{table}

\subsubsection{Experiments on Hard Tasks in DMControl.}
To further evaluate the performance potential of \ours, we conducted additional experiments on three representative hard tasks in DMControl: Hopper, Hop, Walker, Run, and Pendulum, Swingup. Specifically, we trained the models for 500,000 steps and evaluated them 10 times independently after training, reporting the mean and standard deviation of episode rewards (see \Cref{tab:result_on_dmc_3}). The most recent SOTA methods for comparison are TACO~\cite{zheng2023texttt}, Flare~\cite{shang2021reinforcement}, MADI~\cite{grooten2024madi} and ResAct~\cite{liu2025visual}. 

Results show that \ours\ consistently outperforms all baselines across all three tasks. Notably, on Pendulum Swingup (a task demanding fine-grained control), \ours\ achieves not only higher average performance but also lower variance, indicating improved policy stability. This demonstrates the effectiveness of our approach in leveraging the enhanced VLM priors to guide exploration, mitigate sparse reward challenges, and perform better state inference under partial observability. Overall, \ours\ exhibits strong performance across diverse and challenging control tasks, further validating its potential as a general-purpose framework for visual reinforcement learning.

\begin{table}[!ht]
\small
\centering
\setlength{\tabcolsep}{4pt}
\begin{tabular}{c|ccc}
\toprule 
\multirow{1}{*}{Models} & Hopper, Hop  & Walker, Run & Pendulum, Swingup  \\
\midrule
TACO & 112 ± 42 & 355 ± 89 & 485 ± 167\\
Flare & 90 ± 55 &	426 ± 33 & 242 ± 152\\
MADI & 80 ± 24 & 382 ± 87 & 372 ± 101\\
ResAct & 99 ± 49 & 467 ± 27 & 618 ± 380\\
\midrule
\ours\ (Ours) & \cellcolor{gray!20}\textbf{188 ± 9} & \cellcolor{gray!20}\textbf{485 ± 25} & \cellcolor{gray!20}\textbf{792 ± 82}\\
\bottomrule
\end{tabular}
\caption{Testing comparison with SOTA methods of hard tasks in DMControl. The best results for each metric are denoted by \raisebox{0.5ex}
{\colorbox{gray!20}{\;}}.}
\label{tab:result_on_dmc_3}
\end{table}	

\subsubsection{Experiments on Parameters.}
First, we conducted experiments with varying values of $\lambda$, and the results are presented in \Cref{tab:lambda}. It can be observed that both excessively large and small $\lambda$ values lead to a slight performance degradation. This is because not all prior actions derived from VLM reasoning are optimal for any states; making an appropriate trade-off between them and the original policy learning function of visual RL is a necessary condition for achieving global optimal performance.

Then, we evaluated \ours\ using different values of $\psi_{0}$, and the results are summarized in \Cref{tab:psi}. Under a fixed training budget of 100K steps, the performance at 
$\psi_{0}$=2500 is comparable to that at $\psi_{0}$=5000, while a notable performance drop occurs at $\psi_{0}$=7500. This trade-off can be explained as follows: (1) A smaller $\psi_{0}$ triggers more frequent fine-tuning, which not only increases computational overhead but also reduces the accumulation window of the cached dataset $\mathcal{D}_{f}$. This limits the coverage of high-quality trajectories discovered during early RL exploration, potentially weakening the effectiveness of the learned priors. (2) Conversely, a larger $\psi_{0}$ reduces the frequency of fine-tuning, slowing down the adaptation of the VLM to the target domain. As a result, the model fails to provide effective knowledge guidance in the early stages of training.

\begin{table}[!ht]
\normalsize 
\centering
\setlength{\tabcolsep}{4pt}
\begin{tabular}{c|ccc}
\toprule 
\multirow{2}{*}{Metric/Parameters}  & \multicolumn{3}{c}{$\lambda$} \\
& 1.0 & 2.0 & 3.0   \\
\midrule
ER$\uparrow$ & 227 ± 131	& \textbf{248 ± 81} & 232 ± 122 \\
\midrule
DD$\uparrow$ & 237 ± 134 & \textbf{259 ± 85} &  245 ± 126 \\
\bottomrule
\end{tabular}
\caption{Testing performance of different $\lambda$ in the \#HW scenario, where $\psi_{0}$=5000. The best result of each metric is bolded.}
\label{tab:lambda}
\end{table}

\begin{table}[!ht]
\small
\centering
\setlength{\tabcolsep}{4pt}
\begin{tabular}{c|cccc}
\toprule 
\multirow{2}{*}{Metric/Parameters}  & \multicolumn{4}{c}{$\psi_{0}$} \\
& 2500 & 5000 & 7500 & 10,000   \\
\midrule
ER$\uparrow$ & 224 ± 69  & \textbf{248 ± 81} & 187 ± 54 &128 ± 90 \\
\midrule
DD$\uparrow$ & 238 ± 72 & \textbf{259 ± 85} & 195 ± 55 &137 ± 92 \\
\bottomrule
\end{tabular}
\caption{Testing performance of different $\psi_{0}$ in the \#HW scenario, where $\lambda$=2.0. The best result of each metric is bolded.}
\label{tab:psi}
\end{table}

\subsubsection{Experiments on Performance for Each Iteration.}
To demonstrate the interaction between VLM and RL, \Cref{tab:iteration_vlm} and \Cref{tab:iteration_rl} respectively show the gradual improvement in the performance of VLM inference and RL strategies across iterations in the HW scenario.

\begin{table}[!ht]
\small
\centering
\setlength{\tabcolsep}{4pt}
\begin{tabular}{c|cccccc}
\toprule 
\multirow{2}{*}{Metric}  & \multicolumn{6}{c}{Iteration} \\
& 0 & 1 & 2 & 3 & 4 & 5   \\
\midrule
ER$\uparrow$ & -13±7  & 14±7 & 20±8 & 24±11 & 55 ± 13 & 97 ± 22\\
\midrule
DD$\uparrow$ & 5±4 & 21±8  & 27±9 & 41±13  & 60 ± 15 & 102 ± 23\\
\bottomrule
\end{tabular}
\caption{Testing performance of VLM inference for each fine-tuning iteration (from 0) in the \#HW scenario.}
\label{tab:iteration_vlm}
\end{table}

\begin{table}[!ht]
\small
\centering
\setlength{\tabcolsep}{4pt}
\begin{tabular}{c|cccccc}
\toprule 
\multirow{2}{*}{Metric}  & \multicolumn{6}{c}{Iteration} \\
& 1 & 2 & 3 & 4 & 5 & final   \\
\midrule
ER$\uparrow$ & 19±15  & 56±48 & 88±80 & 139±80 & 206±79 & 248±81\\
\midrule
DD$\uparrow$ & 22±15 & 66±54  & 96±82 & 148±84  & 220±83 & 259±85\\
\bottomrule
\end{tabular}
\caption{Testing performance of RL policy after each fine-tuning iteration (from 1) in the \#HW scenario.}
\label{tab:iteration_rl}
\end{table}

\subsubsection{Experiments on different VLMs.}
We have supplemented the fine-tuning effects of different VLMs, including Qwen2-VL-2B~\cite{wang2024qwen2}, Qwen2.5-VL-3B~\cite{bai2025qwen2} and LLava-1.5-7B~\cite{liu2023visual}. The results are shown in \Cref{tab:VLM}. It can be seen that \ours\ performs best under Qwen2.5-VL-3B~\cite{bai2025qwen2}. This might be because the latest Qwen2.5-VL-3B model has the best basic understanding of autonomous driving.

\begin{table}[!ht]
\normalsize
\centering
\setlength{\tabcolsep}{4pt}
\begin{tabular}{c|c|c}
\toprule 
Model  & ER$\uparrow$ &  DD$\uparrow$ \\
\midrule
Qwen2-VL-2B  & 236 ± 69 & 246 ± 72 \\
\midrule
Qwen2.5-VL-3B  & \textbf{248 ± 81} & \textbf{259 ± 85} \\
\midrule
LLava-1.5-7B  & 228 ± 115 & 244 ± 116 \\
\bottomrule
\end{tabular}
\caption{Testing performance of different VLM in the \#HW scenario. The best result of each metric is bolded.}
\label{tab:VLM}
\end{table}

\subsubsection{Experiments on Computational Efficiency.}
We evaluate the GPU memory usage and inference time (IT) of different models during deployment, with results summarized in \Cref{tab:timecop}. As shown, the large VLM model consumes significantly more memory and time compared to the lightweight policy network ($\pi_\theta$). This highlights the efficiency advantage of \ours\ over prior approaches that rely on fine-tuning policies directly on VLMs during inference~\cite{gaven2024sac, zhai2024fine, tan2024true, wei2025gtr}.

During training, we utilize the VLM as an upper-level knowledge source to guide policy learning. Although this introduces additional overhead from VLM fine-tuning, it leads to substantial performance gains in the lightweight policy network, making the trade-off well justified.

\begin{table}[!ht]
\normalsize
\centering
\setlength{\tabcolsep}{4pt}
\begin{tabular}{c|c|c}
\toprule 
Model  & Memory (M) &  Avg. IT (second) \\
\midrule
Qwen2.5-VL-3B  & $\approx $ 8344  &  4.4409 ± 0.7023 \\
\midrule
Qwen2-VL-2B  & $\approx $ 7060 &  4.9421 ± 0.6146 \\
\midrule
LLava-1.5-7B  & $\approx $ 14722 &  3.6634 ± 1.0392 \\
\midrule
$\pi_\theta$ of visual RL & $\approx $ 10 &  0.0012 ± 0.0003\\
\bottomrule
\end{tabular}
\caption{The resource consumption of different models. We tested the average of 10 consecutive frames.}
\label{tab:timecop}
\end{table}

\subsubsection{The Generalization Ability of Cross-scenario.}
To evaluate the generalization capability of cross-scenario policy guidance using VLM fine-tuned by \ours, we designed two test scenarios based on the original highway (\#HW) setting: (1) Scenario A: We modified the route in the \#HW scene by changing both the start and end points. The maximum number of vehicles was reduced from 10 to 5, while maintaining the same weather conditions as in \#HW. (2) Scenario B: We altered the environmental appearance by switching the weather to rainy, thereby introducing visual differences and testing the model’s ability to generalize under varying perceptual conditions.

The results are summarized in \Cref{tab:cross-scenario}. Here, the basic VLM refers to the pre-trained VLM before fine-tuning, while Fine-tuned VLM denotes the VLM after being trained via \ours\ in the HW scenario. As shown, both the VBE and DRL methods benefit from using the fine-tuned VLM, indicating that \ours\ effectively enhances the VLM's ability to provide meaningful policy guidance across different scenes. However, since our current approach does not explicitly model continuous observation sequences, the fine-tuned VLM may still encounter challenges in handling certain dynamic visual inputs. Nevertheless, the observed performance gains in DRL+Fine-tuned VLM methods demonstrate the effectiveness of \ours\ in improving cross-scenario guidance in visual RL.

\begin{table*}[!ht]
\normalsize
\centering
\setlength{\tabcolsep}{4pt}
\begin{tabular}{c|c|cc|cc}
\toprule 
\multirow{2}{*}{Type} & \multirow{2}{*}{Methods} & \multicolumn{2}{c|}{Scenario A} &  \multicolumn{2}{c}{Scenario B}  \\
  &   & ER $\uparrow$ &  DD $\uparrow$  & ER $\uparrow$  & DD $\uparrow$  \\
\midrule
\multirow{2}{*}{Vanilla visual RL}
& SAC &  60 ± 57 & 77 ± 67	& 62 ±42 & 74 ±47 \\
& DeepMDP & 186 ± 100 & 200 ±103	& 139 ± 79 &	153 ± 83\\
\midrule
\multirow{2}{*}{Only VLM}
& VBE+Basic VLM  & -10 ± 6 & 11 ± 3	& -15 ± 10 &	10 ± 6 \\
& VBE+Fine-tuned VLM  & 47 ± 29 & 51 ± 30 & 44 ± 23 & 49 ± 24 \\
\midrule
\multirow{2}{*}{\makecell{VLM-assisted \\ visual RL}} 
& DRL+Basic VLM  & 106 ± 68 & 120 ± 74 & 110 ± 54	& 116 ± 55 \\
& DRL+Fine-tuned VLM & \textbf{251 ± 112} & \textbf{266 ± 114} & \textbf{183 ± 96}	 & \textbf{197 ± 100} \\
\bottomrule
\end{tabular}
\caption{Testing performance about the generalization ability of cross-scenario guidance. The best result of each metric is bolded.}
\label{tab:cross-scenario}
\end{table*}

\begin{figure}[!ht]
\centering
\begin{subfigure}{1\linewidth}
    \includegraphics[width=0.9\linewidth]{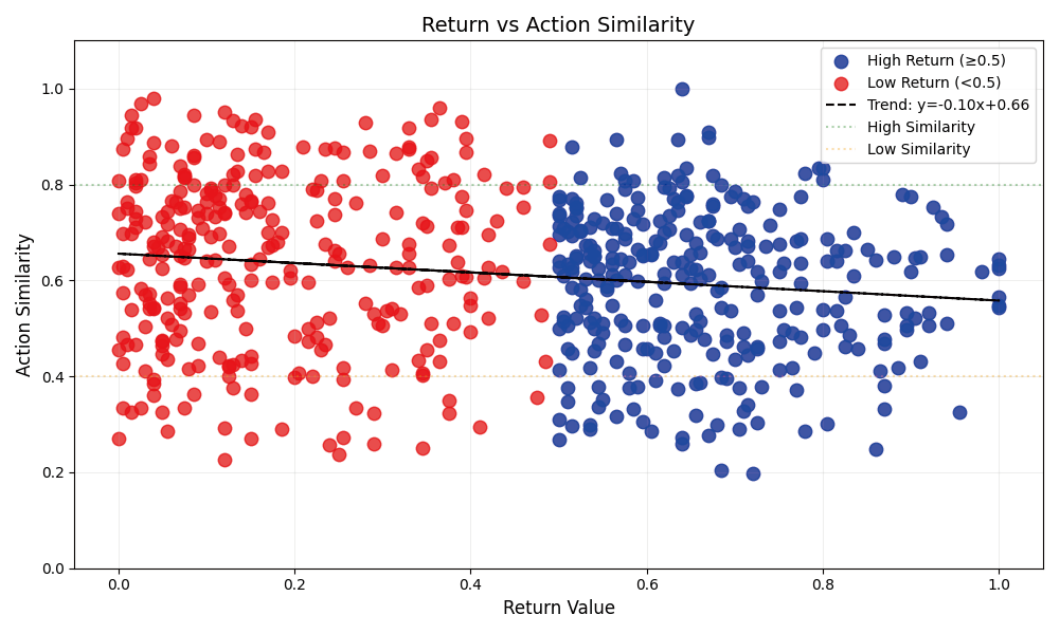}
    \caption{The result of the basic VLM.}
\end{subfigure}
\begin{subfigure}{1\linewidth}
    \includegraphics[width=0.9\linewidth]{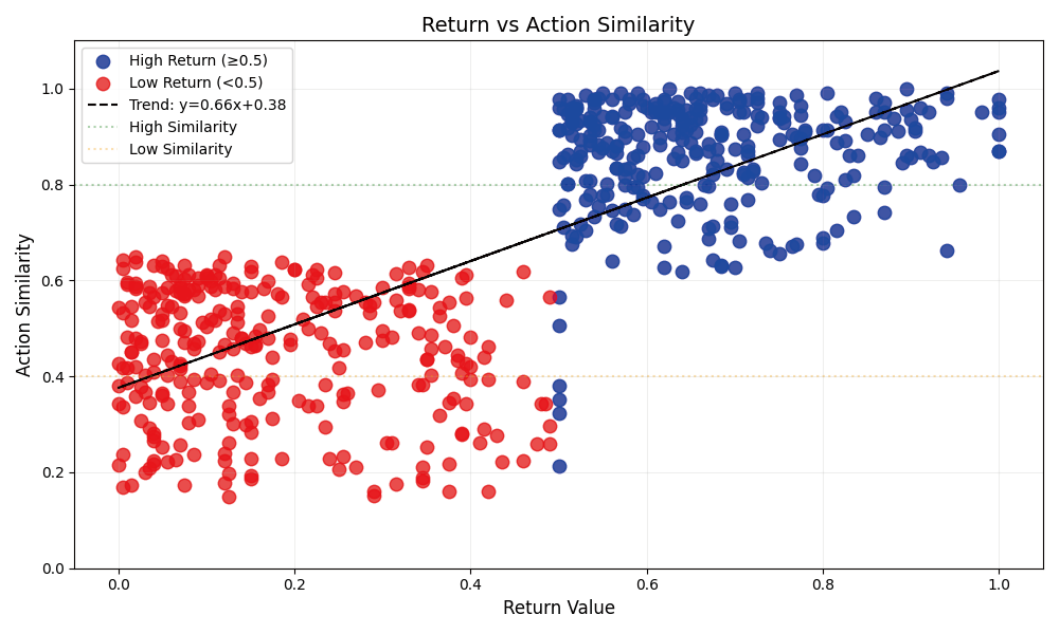}
    \caption{The result of fine-tuned VLM using \ours.}
\end{subfigure}
\caption{Similarity distributions between predicted action and corresponding grounding truth, comparing results before and after the application of the proposed \ours.}
\label{fig:similarity}
\end{figure}

\begin{figure}[!ht] 
  \centering
   \includegraphics[width=0.9\linewidth]{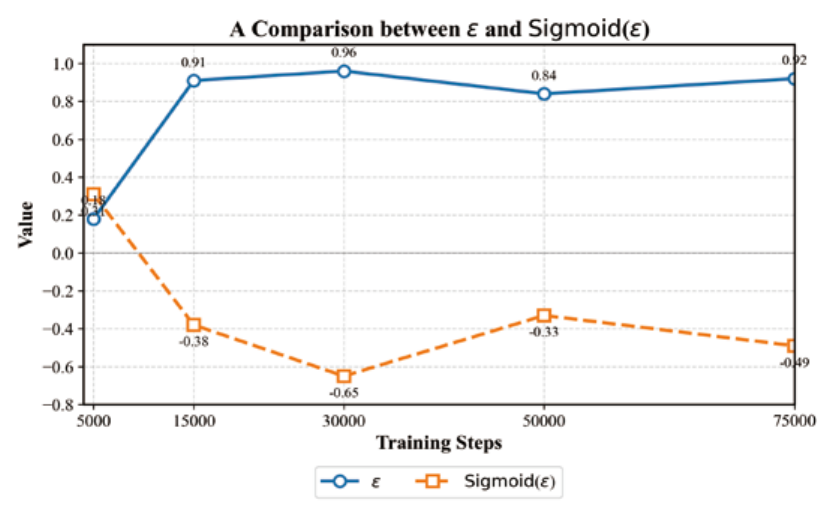}
   \caption{The value variation relationship diagram of $\varepsilon$ and $\mathrm{Sigmoid}(\varepsilon)$ in \Cref{eq:5}.}
   \label{fig:vis_e}
\end{figure}

\begin{figure*}[!ht]
\centering
\begin{subfigure}{0.45\linewidth}
    \includegraphics[width=\linewidth]{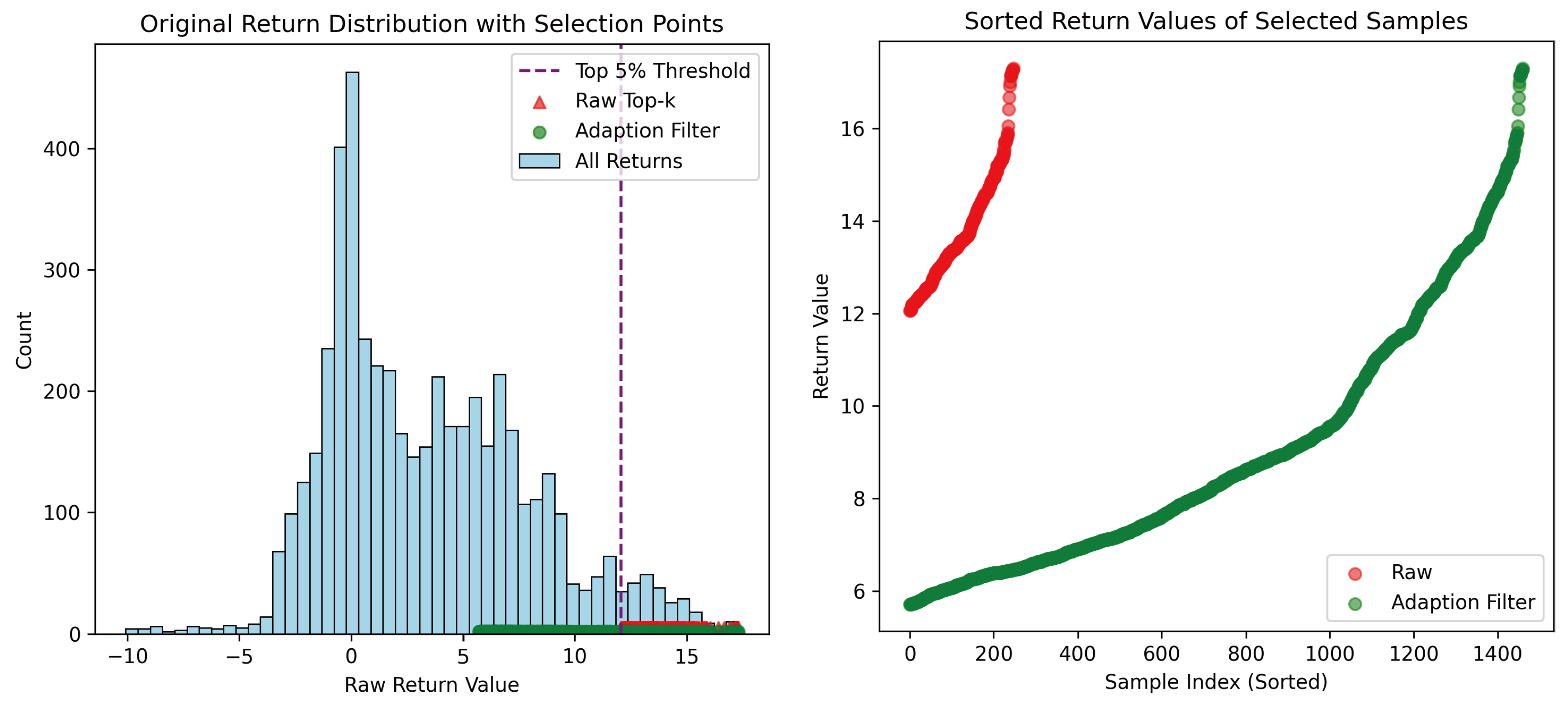}
    \caption{Fine-tuning times $c$=1.}
\end{subfigure}
\begin{subfigure}{0.45\linewidth}
    \includegraphics[width=\linewidth]{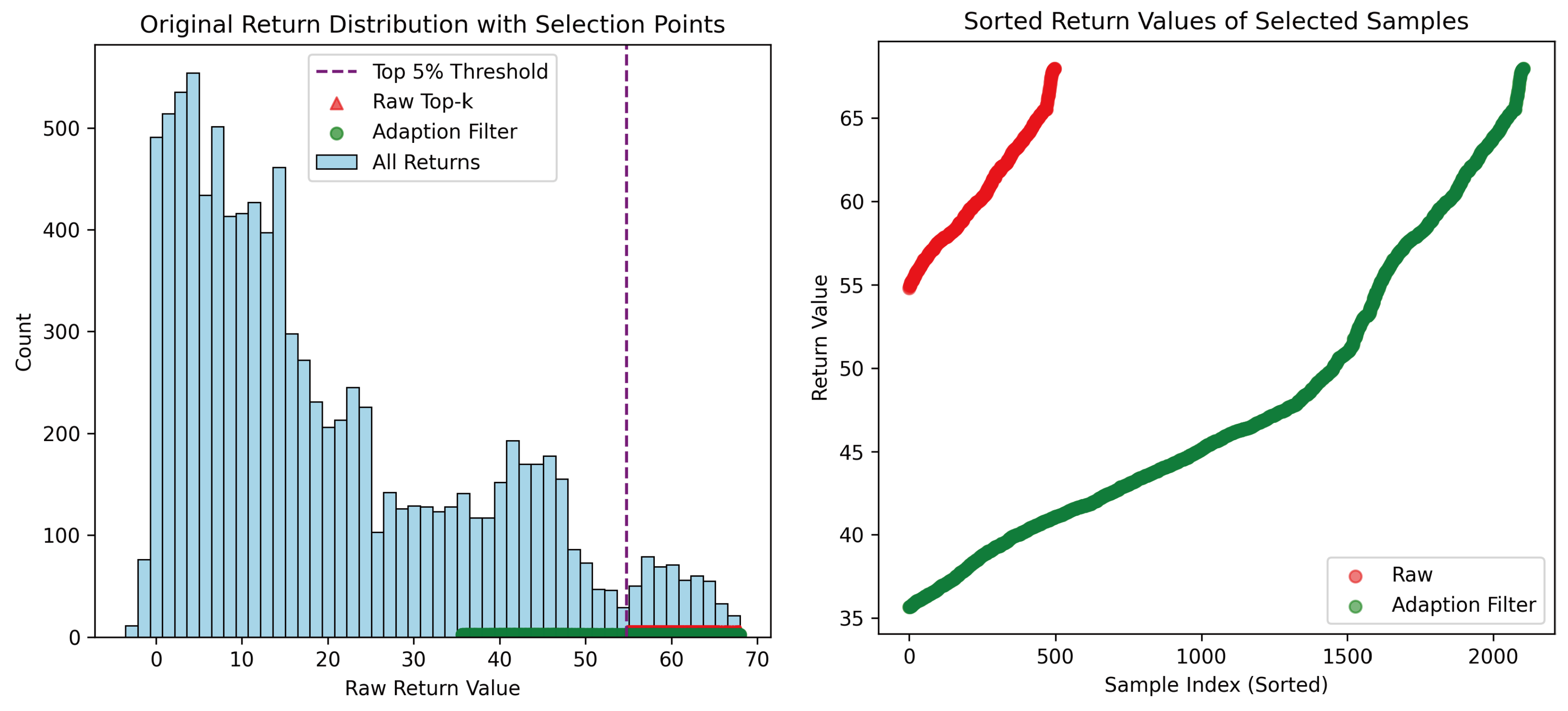}
    \caption{Fine-tuning times $c$=2.}
\end{subfigure}
\begin{subfigure}{0.45\linewidth}
    \includegraphics[width=\linewidth]{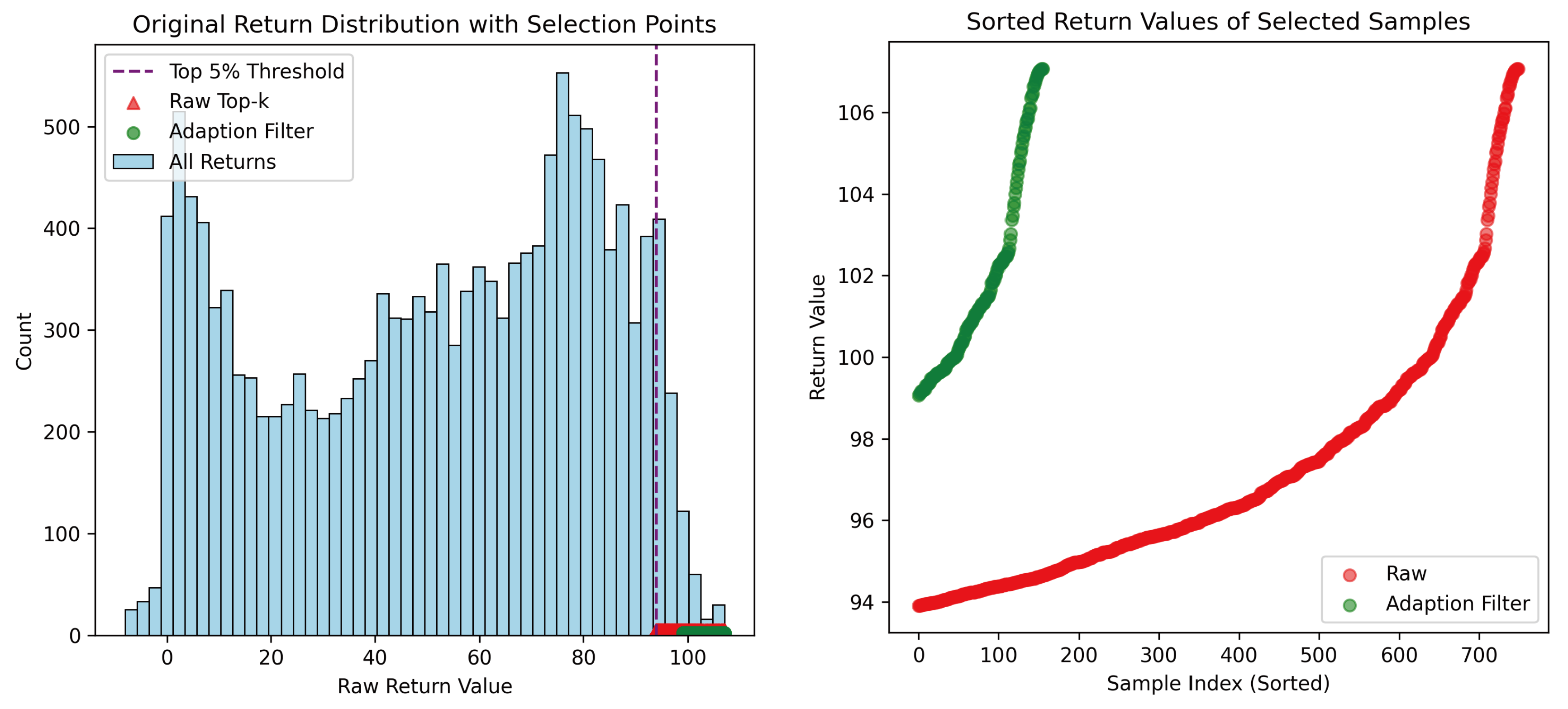}
    \caption{Fine-tuning times $c$=3.}
\end{subfigure}
\begin{subfigure}{0.45\linewidth}
    \includegraphics[width=\linewidth]{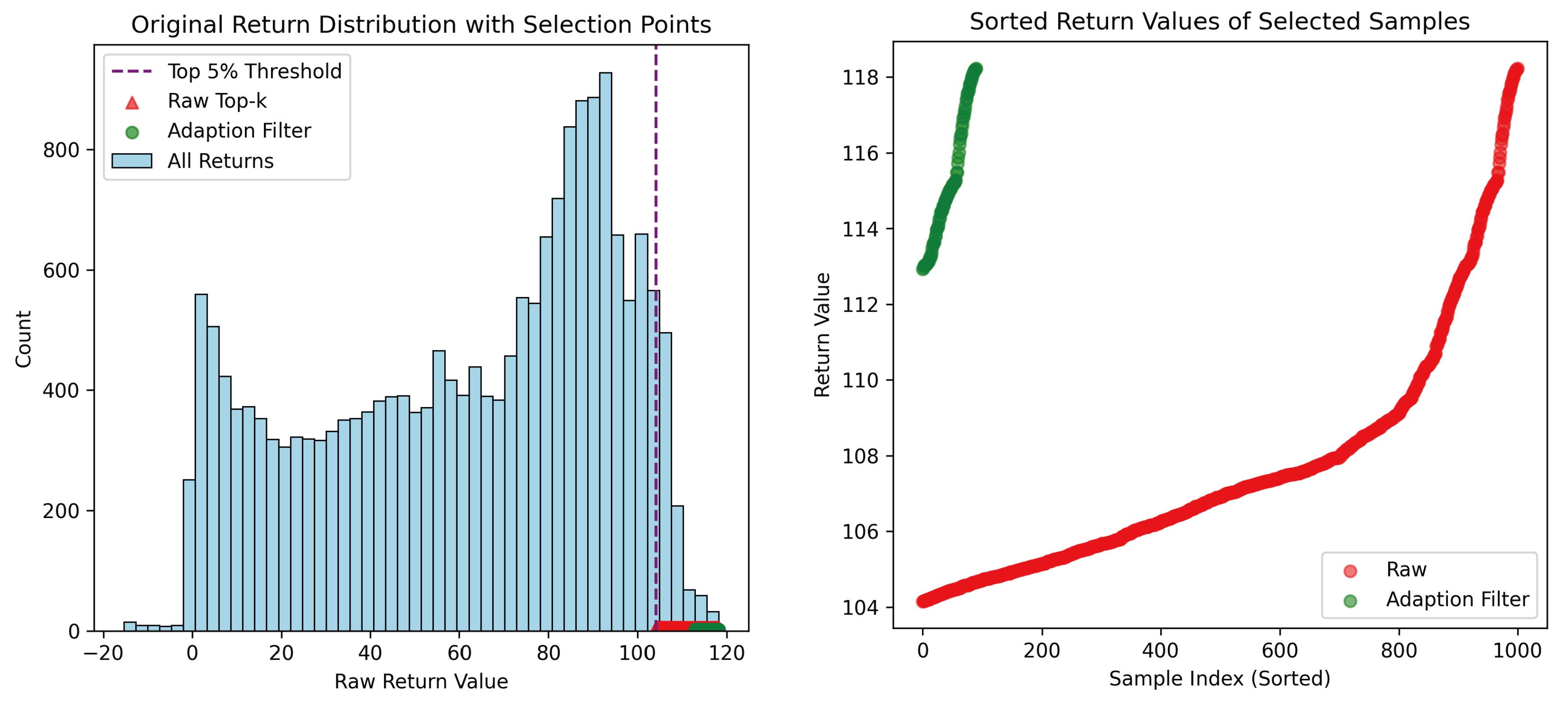}
    \caption{Fine-tuning times $c$=4.}
\end{subfigure}
\caption{A visualization showing (1) the distribution of return data (left figure) in $\mathcal{D}_{f}$ under different fine-tuning times $c$, and (2) the variation in the amount of data filtered by the EDDF module, compared to a baseline that selects the top 95\% samples (right figure).}
\label{fig:vis_c}
\end{figure*}

\begin{figure*}[!ht]
\centering
\begin{subfigure}{1\linewidth}
    \includegraphics[width=\linewidth]{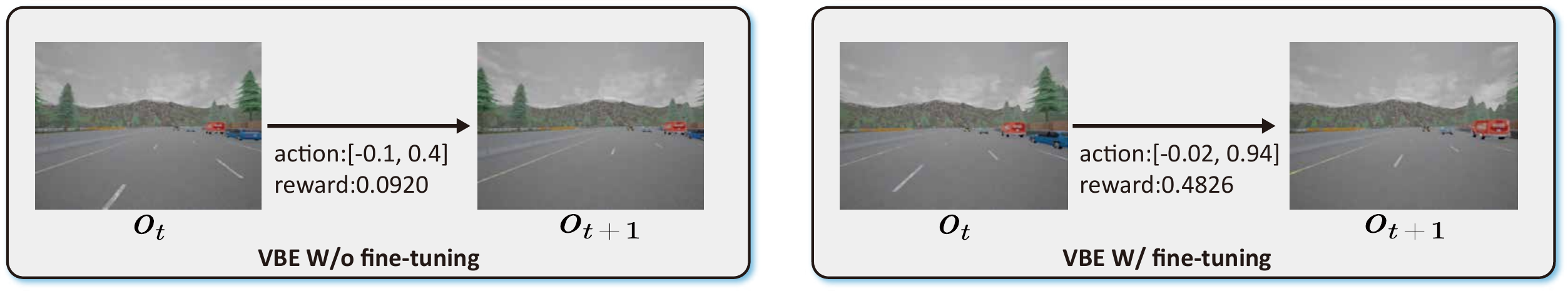}
    \caption{Action and reward outputs For \#HW in CARLA.}
\end{subfigure}
\begin{subfigure}{1\linewidth}
    \includegraphics[width=\linewidth]{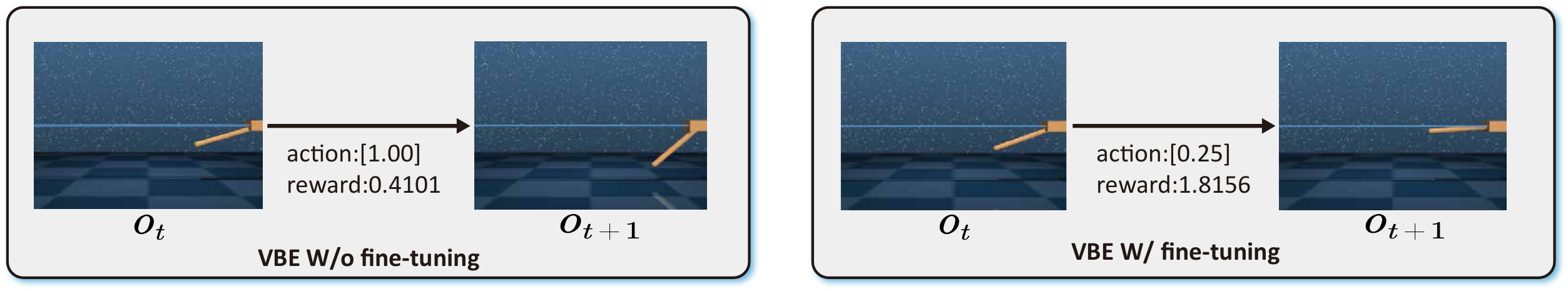}
    \caption{Action and reward outputs For Cartpole, Swingup in DMControl.}
\end{subfigure}
\begin{subfigure}{1\linewidth}
    \includegraphics[width=\linewidth]{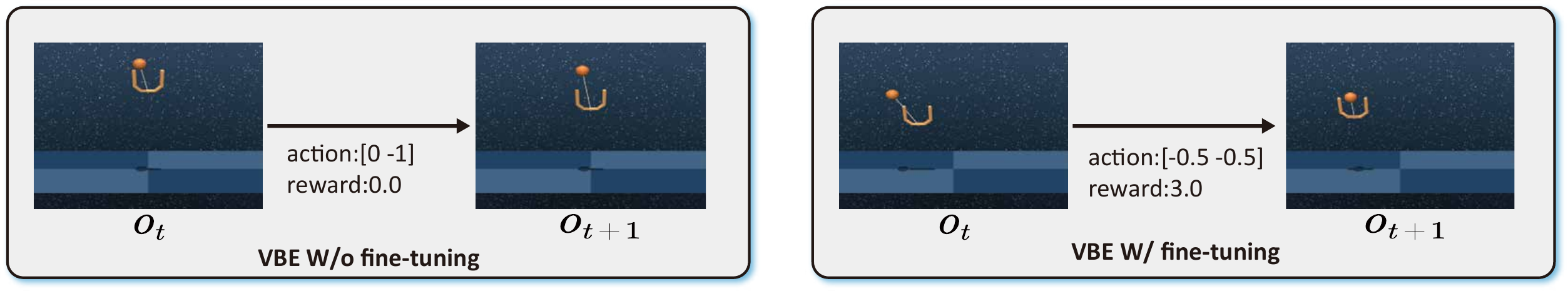}
    \caption{Action and reward outputs For Ball in cup, Catch in DMControl.}
\end{subfigure}
\begin{subfigure}{1\linewidth}
    \includegraphics[width=\linewidth]{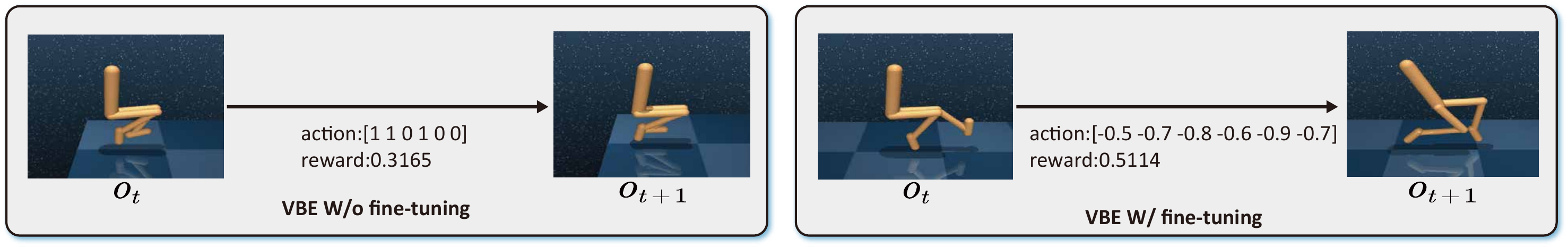}
    \caption{Action and reward outputs For Walker, Walk in DMControl.}
\end{subfigure}
\begin{subfigure}{1\linewidth}
    \includegraphics[width=\linewidth]{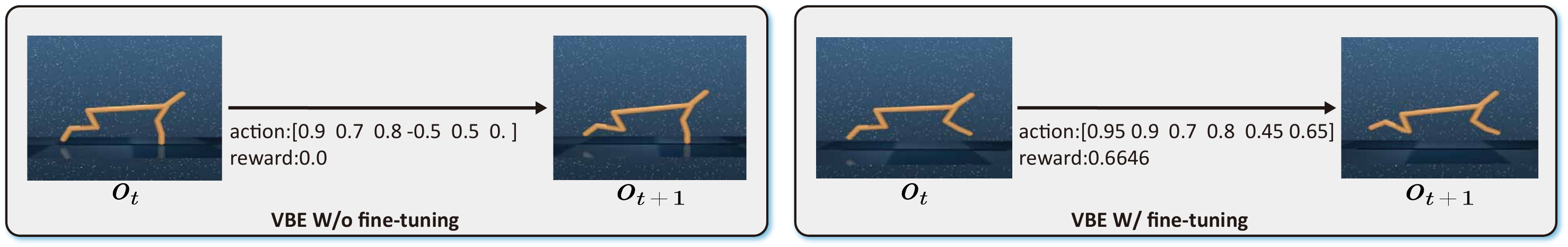}
    \caption{Action and reward outputs For Cheetah, Run in DMControl.}
\end{subfigure}

\caption{Visualization of action and reward outputs before and after VLM fine-tuning. In each subplot, the left panel shows actions from VBE inference using the pretrained VLM, and the right panel shows those using the fine-tuned VLM.}
\label{fig:vbe}
\end{figure*}

\subsection{Visual Analysis}
\subsubsection{Visual Analysis on Reasoning Differences.}
To visually illustrate the differences in reasoning between the basic VLM and the fine-tuned VLM using \ours, we randomly sampled 300 data instances from the trained dataset and separately fed them into both models to generate reasoning-based actions. These predicted actions were then compared with the ground truth using similarity matching, where string outputs are first converted into numerical values and then evaluated via Euclidean distance. The resulting similarities for each sample are visualized in \Cref{fig:similarity}. From the results, we observe the following: 

(1) As shown in \Cref{fig:similarity} (a), the basic VLM (without \ours) fails to reason effectively about high-return actions and does not appropriately suppress low-return actions, leading to random similarity scores.

(2) In contrast, as seen in \Cref{fig:similarity} (b), the fine-tuned VLM under \ours\ shows improved performance. On one hand, high-return reasoning actions align more closely with the true values. On the other hand, in low-return observations, the enhanced generalization of the VLM allows it to reinforce previously inferred actions, resulting in lower similarity to the original low-return values. For instance, in a straight-path observation, the true value for a low-return action is [0.4, 0.3], whereas the fine-tuned VLM predicts [0.01, 0.98].

These findings indicate that \ours\ enables the VLM not only to learn accurate high-return reasoning but also to generalize from low-return observations, thereby improving overall policy guidance in visual reinforcement learning.

\subsubsection{Visual Analysis on EDDF Module.}
\Cref{fig:vis_e} visually illustrates the relationship between $\varepsilon$ and $\mathrm{Sigmoid}(\varepsilon)$ in \Cref{eq:5} during a certain training process. As shown, when $\varepsilon$ is large, $\mathrm{Sigmoid}(\varepsilon)$ tends to be small, leading to a higher threshold $\tau$ and thus more data being filtered. In addition, \Cref{fig:vis_c} illustrates the return distribution in $\mathcal{D}_{f}$ (with $\psi_{0}$=5000) and the variation in filtered data volume using the EDDF module, compared to a top 95\% filtering baseline, across different fine-tuning times $c$. At $c$=1 and $c$=2, the model exhibits high exploration randomness, and the low screening threshold prevents the rejection of potentially valuable low-return samples. As training progresses, the visual RL policy becomes more deterministic (with reduced entropy), leading to an increasing threshold for stricter data filtering. In contrast, the top 95\% method applies a fixed ratio and lacks adaptability to dynamic environmental changes.

\begin{figure*}[!ht] 
  \centering
   \includegraphics[width=1.0\linewidth]{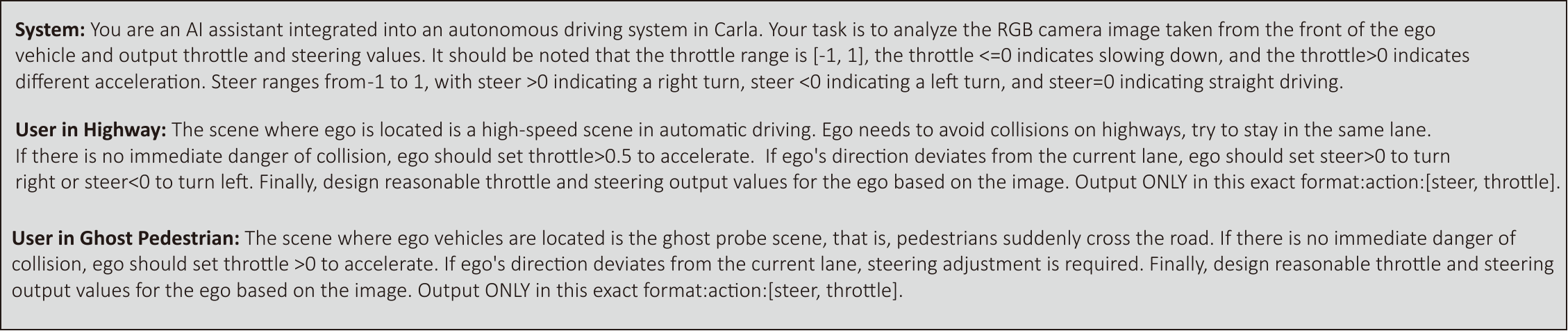}
   \caption{The prompts used in the CARLA benchmarks. }
   \label{fig:prompt}
\end{figure*}

\subsubsection{Visualization of Action Changes Before and After Fine-tuning.}
In \Cref{fig:vbe}, we visualize the differences in VLM inference actions before and after fine-tuning through \ours\ across multiple tasks, along with corresponding rewards. Using VBE (VLM Behavior Executor) for environment interaction, we select trajectory segments with similar (though not identical due to environmental stochasticity) observations to highlight improvements of action changes. The results show that, across high-dimensional state spaces (CARLA), high-dimensional action spaces (Walker and Cheetah), sparse-reward tasks (Ball in cup), and simple control tasks (Cartpole), the fine-tuned VLM generates better actions under similar observations, leading to higher returns. This further illustrates that \ours\ can improve inference robustness in diverse observational conditions, showcasing its effectiveness in integrating VLM priors with visual RL dynamics.

\subsection{Details of the Designed Prompts}
As illustrated in \Cref{fig:prompt}, we provide the prompts (including system and user instructions) used in the CARLA environment. The VLM generates continuous action semantics through reasoning over the input images and prompts, which are then parsed into precise action values, rounded to two decimal places.

\subsection{Limitations and Future Work}
This study has several limitations that point to promising directions for future research:

(1) Currently, \ours\ only generates action semantics during inference and distills this knowledge into the policy network, without leveraging the internal reasoning mechanisms of the VLM. Future work could explore 
incorporating intermediate reasoning paths within the VLM to enhance model interpretability and decision transparency.

(2) Due to computational constraints, the largest VLM used in our experiments is Qwen2.5-VL-3B. However, larger VLMs may provide richer semantic priors that can improve cross-scenario generalization. We plan to evaluate \ours\ with more powerful VLM backbones in future work.

(3) The VLM effect fine-tuned with RL data is not perfect in \ours\ because the RL data itself is noisy and imperfect. Although the performance of the fine-tuned VLM can be improved, it does not fully align with the principles of continuous control. Future work could further focus on fine-tuning to enable VLMs to develop a more comprehensive understanding of environmental knowledge and enhance their temporal reasoning capabilities.

\end{document}